\documentclass[lettersize,journal]{IEEEtran}
\usepackage{amsmath,amsfonts}
\usepackage{algorithmic}
\usepackage{array}
\usepackage[caption=false,font=normalsize,labelfont=sf,textfont=sf]{subfig}
\usepackage{textcomp}
\usepackage{stfloats}
\usepackage{url}
\usepackage{verbatim}
\usepackage{graphicx}
\def\BibTeX{{\rm B\kern-.05em{\sc i\kern-.025em b}\kern-.08em
    T\kern-.1667em\lower.7ex\hbox{E}\kern-.125emX}}
\usepackage{balance}

\usepackage[utf8]{inputenc} 
\usepackage[T1]{fontenc}    
\usepackage{booktabs}       
\usepackage{tabularx}       
\usepackage{multirow}       
\usepackage{adjustbox}
\usepackage{array}
\usepackage{caption}
\usepackage{subfig}

\usepackage{cite}

\begin{document}


\title{HMCGeo: IP Region Prediction Based on Hierarchical Multi-label Classification}

\author{
    \IEEEauthorblockN{Tianzi Zhao\IEEEauthorrefmark{1}, Xinran Liu\IEEEauthorrefmark{2}, Zhaoxin Zhang\IEEEauthorrefmark{1}\thanks{*Zhaoxin Zhang is the corresponding author. This research was supported by the National Key R\&D Program of China (2024QY1103, 2018YFB18002).}, Dong Zhao\IEEEauthorrefmark{3}, Ning Li\IEEEauthorrefmark{1}, Zhichao Zhang\IEEEauthorrefmark{1}, and Xinye Wang\IEEEauthorrefmark{1}}
    
    \IEEEauthorblockA{\IEEEauthorrefmark{1}School of Computer Science and Technology, Harbin Institute of Technology, Harbin, China\\
    Emails: \{23b903088, zhangzhaoxin, 22s030153, li.ning, 22b303010\}@stu.hit.edu.cn}
    
    \IEEEauthorblockA{\IEEEauthorrefmark{2}School of Information and Communication Engineering, Beijing University of Posts and Telecommunications, Beijing, China\\
    Email: xinran\_Liu@bupt.edu.cn}
    
    \IEEEauthorblockA{\IEEEauthorrefmark{3}Tianhe Cyberspace Security Technology Research Institute Co., Ltd., Weihai, China\\
    Email: zhaod@skyvault.cn}
    
}

\markboth{XXXX}%
{Tianzi Zhao \MakeLowercase{\textit{et al.}}: HMCGeo: IP Region Prediction Based on Hierarchical Multi-label Classification}

\maketitle

\bibliographystyle{IEEEtran} 

\bibliographystyle{unsrt} 

\begin{abstract}
Fine-grained IP geolocation plays a critical role in applications such as location-based services and cybersecurity. Most existing fine-grained IP geolocation methods are regression-based; however, due to noise in the input data, these methods typically encounter kilometer-level prediction errors and provide incorrect region information for users. To address this issue, this paper proposes a novel hierarchical multi-label classification framework for IP region prediction, named HMCGeo. This framework treats IP geolocation as a hierarchical multi-label classification problem and employs residual connection-based feature extraction and attention prediction units to predict the target host region across multiple geographical granularities. Furthermore, we introduce probabilistic classification loss during training, combining it with hierarchical cross-entropy loss to form a composite loss function. This approach optimizes predictions by utilizing hierarchical constraints between regions at different granularities. IP region prediction experiments on the New York, Los Angeles, and Shanghai datasets demonstrate that HMCGeo achieves superior performance across all geographical granularities, significantly outperforming existing IP geolocation methods.
\end{abstract}


\section{Introduction}
\IEEEPARstart{I}{P} geolocation is a technique used to predict the geographical location of a host based on its IP address \cite{1}, playing a crucial role in location-based services, network topology optimization, and cybersecurity \cite{35}, \cite{2}, \cite{3}, \cite{4}, \cite{5}, \cite{6}, \cite{7}. Using IP geolocation technology, online services and applications infer the geographical location of users to deliver localized weather updates, news, and event notifications \cite{2}. Internet service providers (ISPs) estimate the approximate location of target hosts to optimize traffic transmission paths, reduce network latency, and improve transmission efficiency \cite{3}. Network analysts examine the geographical origins of incoming traffic to assess security threats from suspicious addresses. This process plays a critical role in mitigating Distributed Denial of Service (DDoS) attacks and assisting in the detection of Border Gateway Protocol (BGP) hijacking incidents \cite{4}. Based on the accuracy of prediction results, IP geolocation is categorized into coarse-grained and fine-grained geolocation. Coarse-grained IP geolocation predicts the location of a target host by utilizing allocation information such as Autonomous System Numbers (ASN), ISP, and BGP, or by analyzing the relationship between latency and distance. These methods construct geolocation databases that provide location information at the country or city level. However, the prediction error typically spans several tens of kilometers. Building on this foundation, fine-grained IP geolocation reduces prediction errors to a few kilometers in certain regions by leveraging richer landmarks or employing more effective prediction methods. As downstream tasks such as precise risk management and smart cities increasingly demand higher geolocation accuracy, current research primarily focuses on fine-grained IP geolocation \cite{6}, \cite{7}. Although current fine-grained IP geolocation methods, such as Checkin-Geo \cite{13}, GeoCam \cite{14}, GNNGeo \cite{18}, and GraphGeo \cite{8}, have achieved excellent results under specific conditions, they still face numerous challenges. \textbf{In this study, we aim to predict the region associated with the target host, providing socially relevant insights for downstream tasks.} 
\begin{figure}[!t]
\centering
\includegraphics[width=8.5cm]{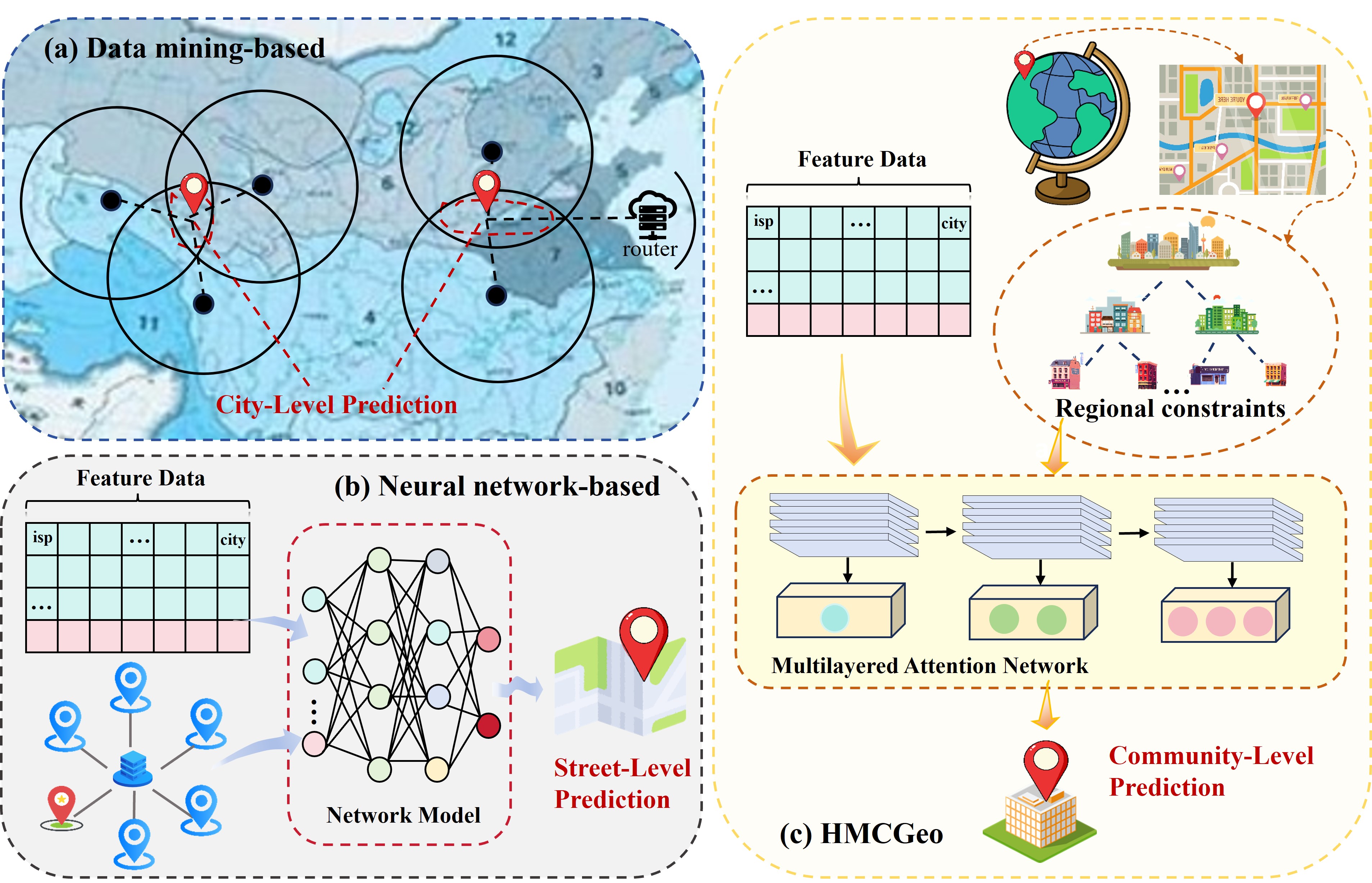}
\caption{Overview of IP Geolocation Methods.}
\label{fig1}
\end{figure}

Fine-grained IP geolocation can be categorized into data mining-based methods and learning-based methods based on their underlying approaches. Data mining-based methods attribute prediction errors mainly to the insufficient availability of reliable landmarks, which are hosts with known IP addresses and geographical locations. These methods focus on extracting trustworthy landmarks from public resources and utilizing them in IP geolocation to enhance prediction accuracy. For example, GeoCam \cite{14} extracts clues from IP cameras, while Checkin-Geo \cite{13} leverages social media check-ins to identify reliable landmarks, enhancing the overall geolocation accuracy. The effectiveness of such methods typically depends on the quality of landmarks, including their availability, accuracy, and spatial distribution. \textbf{This reliance limits their performance in regions with poor landmark quality or underdeveloped network resources.}

Learning-based methods suggest that algorithm design also influences prediction errors. These methods often employ techniques such as crowdsourcing or manual data collection to gather sufficient landmarks in designated regions, with a primary focus on enhancing the performance of prediction algorithms. With the rapid development of artificial intelligence technologies, numerous deep learning methods for IP geolocation have emerged in recent years. These data-driven methods demonstrate strong representational capabilities, effectively learning the complex nonlinear relationships between IP features such as attribute and measurement features, and the geographical locations of their hosts. For example, GNNGeo \cite{18} and GraphGeo \cite{8} leverage graph neural networks to model fine-grained IP geolocation as a regression problem on attributed graphs. By considering the attributes of neighboring hosts in the graph, these methods enable more comprehensive feature extraction, significantly improving prediction performance. Although these methods have demonstrated outstanding performance in geolocation tasks, they still face significant challenges. Currently, most of these approaches treat fine-grained IP geolocation as a regression task, attempting to learn an accurate "feature-coordinate" mapping through complex network architectures to predict the coordinates of the target host. However, they neglect the accuracy of the region associated with the target host \cite{8}. Studies reveal that a significant amount of noise inherent in IP features is difficult to eliminate, resulting in prediction error remaining at the kilometer scale \cite{17}, \cite{18}, \cite{19}, \cite{20}. This limitation makes such methods prone to providing incorrect region information. \textbf{Furthermore, their disregard for social attributes does not meet the requirements of downstream tasks, such as criminal tracking.}

In this study, to address the challenges, we propose a framework named HMCGeo, which focuses on IP region prediction and aims to predict the region of an IP across multiple geographical granularities. The social attributes inherent in geographical regions, such as streets and communities, can be crucial for downstream tasks. However, as revealed in our investigation (Section \ref{Analysis of IP Geolocation Performance} and \ref{Case Analysis}), existing methods perform poorly in region prediction tasks. We analyze the distribution patterns of IP hosts and find that hosts within the same geographical region often exhibit similar patterns. Inspired by this observation, we partition cities into hierarchical granularities such as "district-street-community" and propose HMCGeo, which models fine-grained IP geolocation as a hierarchical multi-label classification task. Specifically, we first map the locations of landmarks to regions at different granularities and establish hierarchical relationships between these regions. Next, we select the corresponding set of landmarks for the target host based on whether they share the same last-hop router. Finally, we extract information from IP features at each granularity step by step and provide prediction results at the respective granularity. By combining local and global outputs, we comprehensively consider IP features from both granular and holistic perspectives. Additionally, we incorporate probabilistic classification loss into HMCGeo to further optimize prediction performance, significantly improving the accuracy of community-level IP geolocation. We conduct experiments on real-world IP geolocation datasets to evaluate the performance of HMCGeo. The results demonstrate that, compared to previous fine-grained IP geolocation methods, HMCGeo achieves significantly higher region prediction accuracy, with improvements ranging from 198.2\% to 546.7\% at the finest geographical granularity. In summary, our contributions can be categorized into the following three aspects.

\textbf{(1) Novel IP Geolocation Perspective:} By analyzing the distribution patterns of IP addresses, we explore the relationship between IP allocation and geographical regions. Inspired by this observation, we propose modeling fine-grained IP geolocation as a hierarchical multi-label classification problem. Unlike existing methods that focus on predicting the precise coordinates of an IP, we argue that concentrating on predicting the region of the target host represents a promising perspective.

\textbf{(2) Innovative IP Regional Prediction Framework:} To address the challenges of fine-grained IP geolocation, we design a hierarchical multi-classification framework named HMCGeo. Falling under the category of learning-based methods, this framework transforms the IP geolocation problem into a classification task through preprocessing steps such as landmark host mapping, hierarchical region extraction, and topology-based landmark selection. HMCGeo employs feature extraction units and attention mechanisms to model the relationships between target and landmark hosts across geographical granularities, delivering accurate predictions. Additionally, it incorporates result fusion and probabilistic classification loss to further enhance prediction performance.

\textbf{(3) Outstanding Regional Prediction Performance:} HMCGeo demonstrates superior regional prediction performance on datasets from New York, Los Angeles, and Shanghai, which include over 300,000 real IP addresses. It significantly outperforms existing methods, with the performance improvement becoming more pronounced as the geographical granularity increases. Specifically, HMCGeo achieves a 1\%-24.8\% improvement at coarse granularity and a 39.0\%-546.7\% improvement at fine granularity. HMCGeo can serve as a complementary approach to existing fine-grained IP geolocation methods, enriching the results with additional socially relevant insights.

The remainder of this paper is organized as follows: Section \ref{RELATED WORK} reviews related work, including IP geolocation and hierarchical multi-label classification. Section \ref{IP Distribution Analysis} discusses the motivation for this study from the perspective of IP allocation patterns. Section \ref{Proposed Methods} defines the problem and provides a detailed description of the HMCGeo framework, covering landmark host mapping and hierarchical region extraction, topology-based landmark selection, hierarchical IP geolocation, and the loss function. Section \ref{Experiments} presents the experimental design and results analysis, evaluating the performance of HMCGeo on regional prediction tasks. Finally, Section \ref{Conclusion} concludes the paper.

\section{Related Work} \label{RELATED WORK}
In this section, we present research work related to HMCGeo, focusing primarily on IP geolocation and hierarchical multi-label classification.
\subsection{IP Geolocation}
Coarse-grained IP geolocation leverages IP attribute and measurement information to estimate the approximate location of target hosts, with typical errors exceeding 10 km \cite{9}, \cite{10}, \cite{11}, \cite{12}. Because it requires minimal reliance on landmarks and is easy to deploy, this approach is widely adopted in various services that offer related query functionalities \cite{45}, \cite{46}, \cite{47}. However, the significant prediction errors make it unsuitable for downstream tasks such as network security. 

By extracting more reliable landmarks or employing more efficient feature extraction components, fine-grained IP geolocation reduces prediction errors in pursuit of street-level accuracy. In this paper, we categorize such methods into data mining-based methods and neural network-based methods. \textbf{Data mining-based methods} extract various landmark clues from internet resources to infer the locations of target hosts, aiming to improve prediction accuracy. For instance, SLG identifies landmarks from online maps and integrates algorithms like CBG to progressively narrow down the target host's region, achieving street-level IP geolocation \cite{6}, \cite{9}. Checkin-Geo gathers log data from social media and analyzes user check-in information to determine the relationship between IP addresses and locations, significantly enhancing geolocation accuracy \cite{13}. Dan et al. generate landmarks by performing reverse DNS queries to extract domain names associated with IP addresses and analyzing embedded location clues, whereas Luo et al. focus on domain names linked to web content and extract location information directly from web pages \cite{15}, \cite{39}, \cite{40}, \cite{41}, \cite{42}. GeoCam and ProbeGeo specialize in using IP cameras and other online IoT devices. GeoCam retrieves publicly available video streams from the internet and extracts location data from corresponding web pages, while ProbeGeo extends this approach to various IoT devices by designing a comprehensive landmark extraction framework for probes and common landmarks \cite{14}, \cite{43}. Although these methods approach landmark extraction from diverse perspectives and significantly enhance IP geolocation performance, they still face notable limitations. For example, SLG and reverse DNS methods can only provide accurate landmarks when web servers are locally deployed. With the rise of content delivery networks (CDNs) and cloud computing, location clues on web pages may no longer correspond to the IP addresses, causing performance degradation. Similarly, Checkin-Geo relies on social media logs, which are often sensitive data controlled by large internet companies and difficult for individuals or organizations to access. Furthermore, the method's applicability is restricted by the geographical reach of the company's services. GeoCam and ProbeGeo depend on publicly accessible IoT devices. However, increasing government regulations and growing privacy awareness have led more network devices to hide their IP addresses, limiting the effectiveness of these methods. For example, ProbeGeo has identified only about 80,000 landmarks globally, with uneven distribution across countries. These constraints restrict the number of usable landmarks, impeding further optimization of IP geolocation performance.

\textbf{Neural network-based methods} typically employ crowdsourcing to collect sufficient landmarks, with a primary focus on improving algorithm performance. These methods leverage neural networks to learn the nonlinear relationship between IP features and their geographical locations, using this knowledge to predict the locations of target hosts. NNGeo is the first to apply neural network architectures to IP geolocation tasks, utilizing a multilayer perceptron (MLP) to learn the mapping between latency measurements and coordinates. Building on this, MLPGeo further incorporates routing information to enhance prediction accuracy \cite{16}, \cite{17}. With advancements in deep learning, various neural network architectures emerge, and researchers find that graph neural networks (GNNs) are particularly well-suited for modeling computer networks. Against this backdrop, Ding et al. propose GNNGeo, which models IP networks as undirected graphs and formulates IP geolocation as an attributed graph node regression problem \cite{36}, \cite{37}. This approach profoundly influences subsequent research \cite{18}. Building on GNNGeo, GraphGeo improves the framework by integrating prior knowledge and topological relationships between hosts to construct high-quality attributed graphs. It also introduces an uncertainty-aware GNN framework to mitigate prediction errors caused by network jitter and congestion \cite{8}. NeighborGeo further optimizes the adjacency graph construction process in GraphGeo, recognizing that the selection of neighboring nodes significantly affects prediction accuracy. By incorporating graph structure learning, it refines neighbor selection and improves prediction performance \cite{44}. Compared to reducing prediction errors, RIPGeo and TrustGeo focus more on the reliability of prediction results. RIPGeo emphasizes the robustness of IP geolocation systems, proposing a perturbation training strategy that introduces noise during training to enhance the stability of graph representation learning \cite{20}. In contrast, TrustGeo prioritizes the trustworthiness of predictions, introducing a reliable street-level IP geolocation framework based on evidential deep learning. This framework parameterizes the mean and variance during the prediction process, providing both prediction results and confidence scores, thereby offering additional data support for developers to monitor system performance \cite{19}. Despite optimizing prediction errors to within hundreds to thousands of meters, advanced fine-grained IP geolocation methods such as GraphGeo, RIPGeo, and TrustGeo still face significant challenges. Existing research shows that noise persists in IP geolocation data due to outdated attribute features and network congestion in measurement environments \cite{20}. These methods primarily aim to learn the mapping between IP features and host coordinates. However, noise often causes predictions to cluster around landmark centers, failing to provide accurate regional information about the target host. To address this limitation, this paper proposes HMCGeo.

\begin{table*}[h]
\centering
\renewcommand{\arraystretch}{1.2} 
\caption{Shanghai IP Landmark Cluster Analysis Results}
\label{tab:IP_cluster_analysis}
\setlength{\tabcolsep}{5pt}
\begin{tabularx}{\textwidth}{>{\centering\arraybackslash}m{2cm} >{\centering\arraybackslash}m{2.7cm} >{\centering\arraybackslash}m{1cm} >{\centering\arraybackslash}m{1cm} >{\centering\arraybackslash}m{1.5cm} >{\centering\arraybackslash}m{3.5cm} >{\centering\arraybackslash}m{3.5cm}}
\toprule
 & & Batchs & IPs & Cluster \% & Avg intra-cluster Dist. (km) & Min Inter-cluster Dist. (km) \\
\midrule
\multirow{3}{*}{Clusters < 2} & Not clustered & 143 & 959 & 0.00 & - & - \\
 & All in one cluster & 134 & 10637 & 1.00 & 0.12 & - \\
 & Partially in one cluster & 155 & 4060 & 0.37 & 0.17 & - \\
Clusters > 1 & - & 562 & 90819 & 0.82 & 0.20 & 2.50 \\
\midrule
Total & & 994 & 106475 & 0.80 & 0.18 & 2.50 \\
\bottomrule
\end{tabularx}
\end{table*}

\subsection{Hierarchical Multi-Label Classification}
Hierarchical Multi-Label Classification (HMC) is a classification task where each sample can simultaneously belong to multiple hierarchically constrained categories. This approach is widely applied in fields such as text classification \cite{21}, image classification \cite{22}, and bioinformatics \cite{23}. HMC methods are generally categorized into local output methods and global output methods based on their output strategies. Local output methods \cite{24} adopt a top-down strategy, making independent predictions at each hierarchical level to ensure high-quality results for the corresponding level. In contrast, global output methods \cite{25} use a single classifier to predict categories across all levels simultaneously. Although global output methods have simpler structures, they are prone to propagation errors and often fail to capture the finer details of hierarchical constraints. On the other hand, local output methods focus exclusively on features at the current level during prediction, which may hinder their ability to effectively model hierarchical relationships between categories. From the perspective of hierarchical constraints, HMC methods can be further classified into post-processing methods and loss function constraint methods. Post-processing methods \cite{26} impose hierarchical constraints on prediction results after the prediction process, ensuring compliance with category relationships at the expense of reduced accuracy. In contrast, loss function constraint methods embed hierarchical constraints directly into the loss function, allowing them to guide the model during training. These methods aim to improve prediction accuracy by leveraging hierarchical constraints without strictly enforcing compliance with category relationships \cite{27}. To achieve the most accurate prediction of the region associated with an IP, this paper adopts the loss function constraint approach. Specifically, we utilize probabilistic classification loss to optimize prediction results for IP region association.

\section{IP Distribution Analysis} \label{IP Distribution Analysis}
Understanding the distribution patterns of IP addresses is crucial for IP geolocation. Taking Shanghai as an example, we conduct a clustering analysis on the locations of over 100,000 hosts to explore IP distribution patterns. The specific setup is as follows: previous studies indicate that hosts sharing the same last-hop router exhibit strong geographical correlations. Based on this observation, we first apply IP host aggregation \cite{7}, grouping all IPs into multiple batches, with each batch containing all hosts under the same last-hop router. We perform clustering analysis on the hosts within each batch using the DBSCAN method \cite{28}. The specific parameter settings are as follows: "eps = 0.3 km" and "min\_samples = 3". The statistical results of the clustering analysis are presented in TABLE \ref{tab:IP_cluster_analysis}.

\begin{figure}[h]
\centering
\subfloat[\scriptsize Not clustered\label{subfig:not_clustered}]{\includegraphics[width=0.16\textwidth]{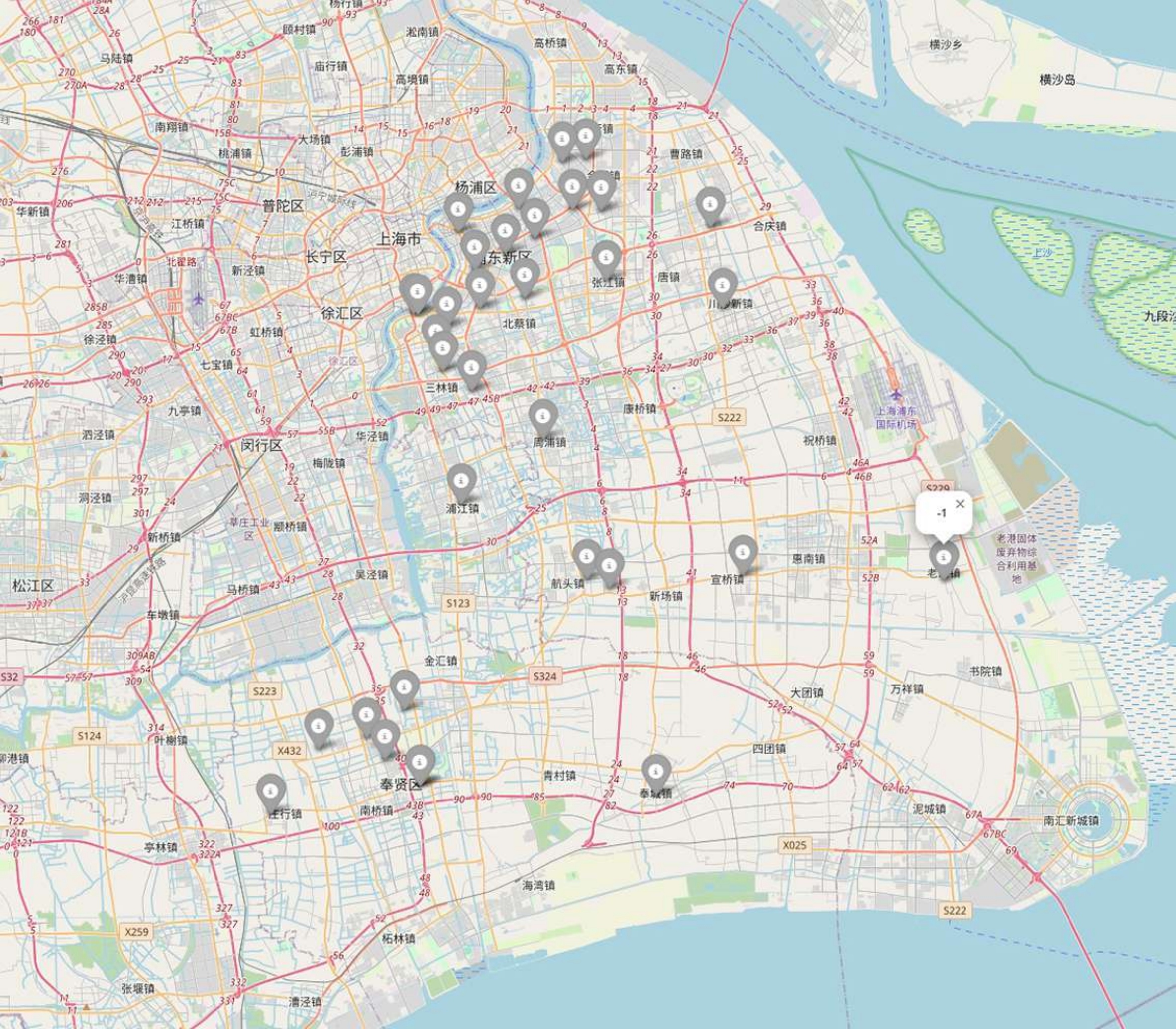}}\hfill
\subfloat[\scriptsize All in one cluster\label{subfig:all_in_one}]{\includegraphics[width=0.16\textwidth]{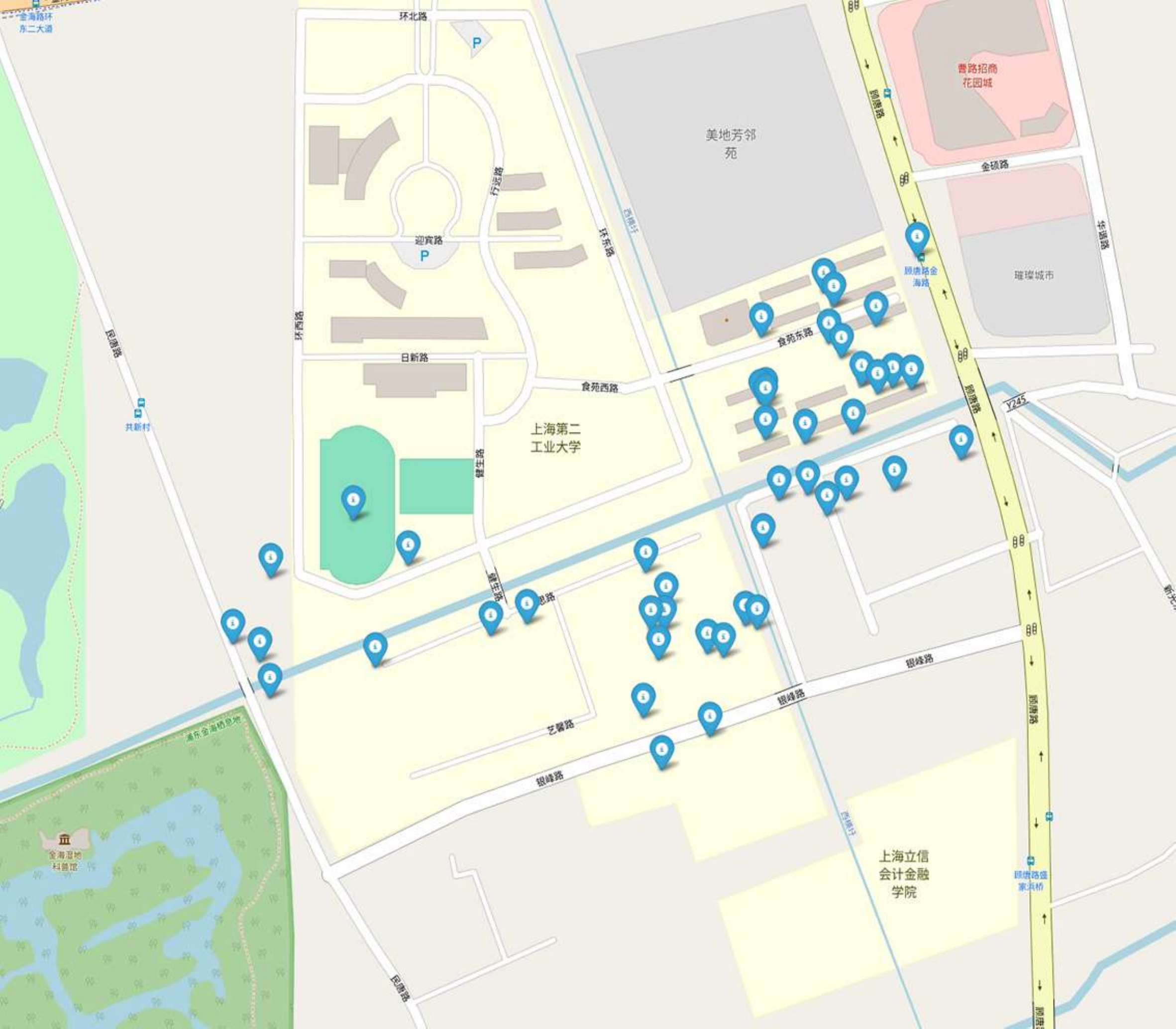}}\hfill
\subfloat[\scriptsize Partially in one cluster\label{subfig:partially_in_one}]{\includegraphics[width=0.16\textwidth]{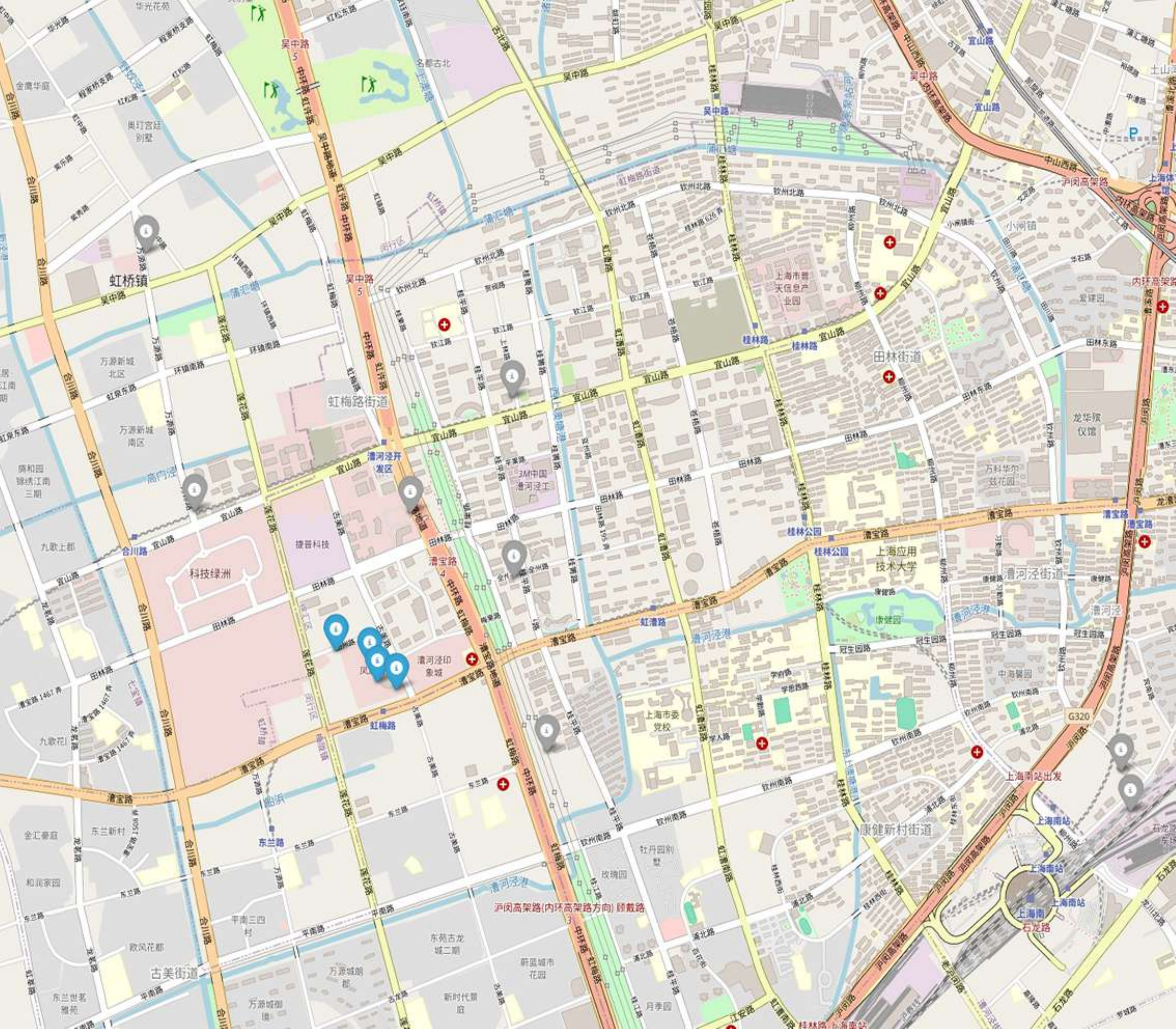}}
\caption{Three typical examples with fewer than 2 clusters}
\label{fig:three_case}
\end{figure}

We observe that the clustering results can be divided into two categories: \textit{clusters $< 2$} and \textit{clusters $\geq 2$}. Hosts in the \textit{clusters $< 2$} category account for less than 15\% of all hosts, while the remaining hosts fall under the \textit{clusters $\geq 2$} category. The \textit{clusters $< 2$} category can be further divided into three subcategories: hosts in the batch are not clustered, all hosts are grouped into a single cluster, or only some hosts are grouped into a single cluster. As shown in Fig.\ref{subfig:not_clustered}, Fig.\ref{subfig:all_in_one}, and Fig.\ref{subfig:partially_in_one}, we provide three examples to illustrate these scenarios.

As shown in Fig.\ref{subfig:not_clustered}, none of the hosts in this batch form clusters. This can be attributed to two main reasons: (1) The batch contains too few hosts to meet the clustering criteria. Among the 143 unclustered batches, 90 have fewer than six hosts. (2) The hosts are widely dispersed. For instance, in Fig.\ref{subfig:not_clustered}, the hosts are scattered across the entire Shanghai urban area, with distances between them exceeding the clustering radius. Such dispersed distributions lack clear patterns, making accurate location prediction challenging. However, hosts in this category account for less than 1\% of the total. Fig.\ref{subfig:all_in_one} depicts a batch where all hosts are grouped into a single cluster. The average intra-cluster distance in these cases is 0.12 km, with most hosts concentrated within a single community, as shown in Fig.\ref{subfig:all_in_one}. Approximately 10\% of hosts fall into this category, making IP geolocation tasks relatively straightforward since a single landmark can accurately determine their locations. This pattern is likely influenced by the IP allocation strategies used by ISPs. Fig.\ref{subfig:partially_in_one} shows a batch where only a portion of the hosts form a single cluster. Among these, 37\% of hosts are clustered with an average intra-cluster distance of 0.17 km. The remaining hosts are unclustered, likely due to the reasons mentioned earlier, such as insufficient landmark data. As seen in Fig.\ref{subfig:partially_in_one}, the unclustered hosts appear isolated. Hosts in this category make up a small fraction of the total, accounting for less than 4\%.

\begin{figure}[h]
\centering
\subfloat[\scriptsize Comparison of inter-cluster and intra-cluster distances\label{subfig:comparison}]{\includegraphics[width=0.24\textwidth]{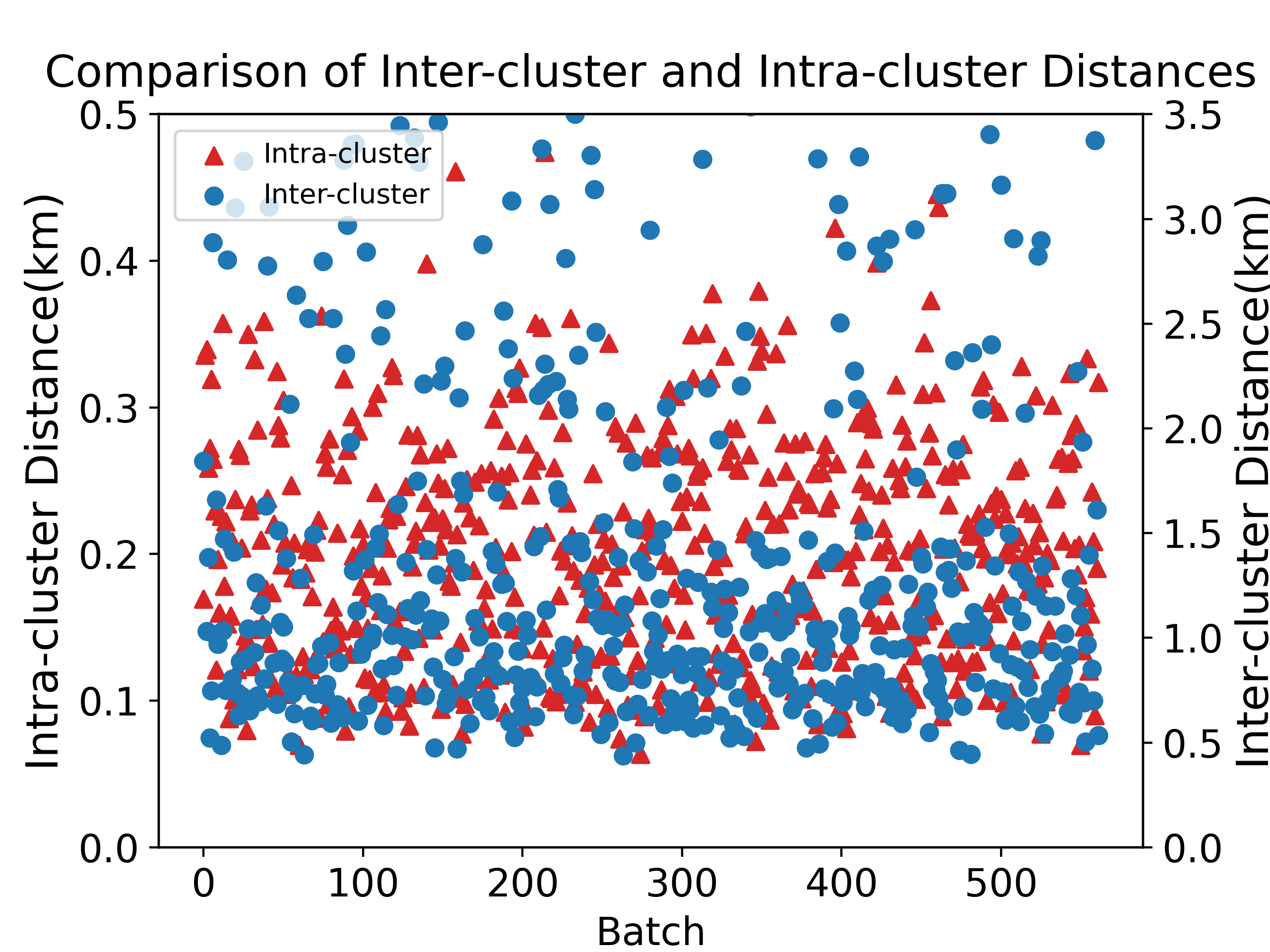}}\hfill
\subfloat[\scriptsize Clustered into multiple groups\label{subfig:multiple}]{\includegraphics[width=0.24\textwidth]{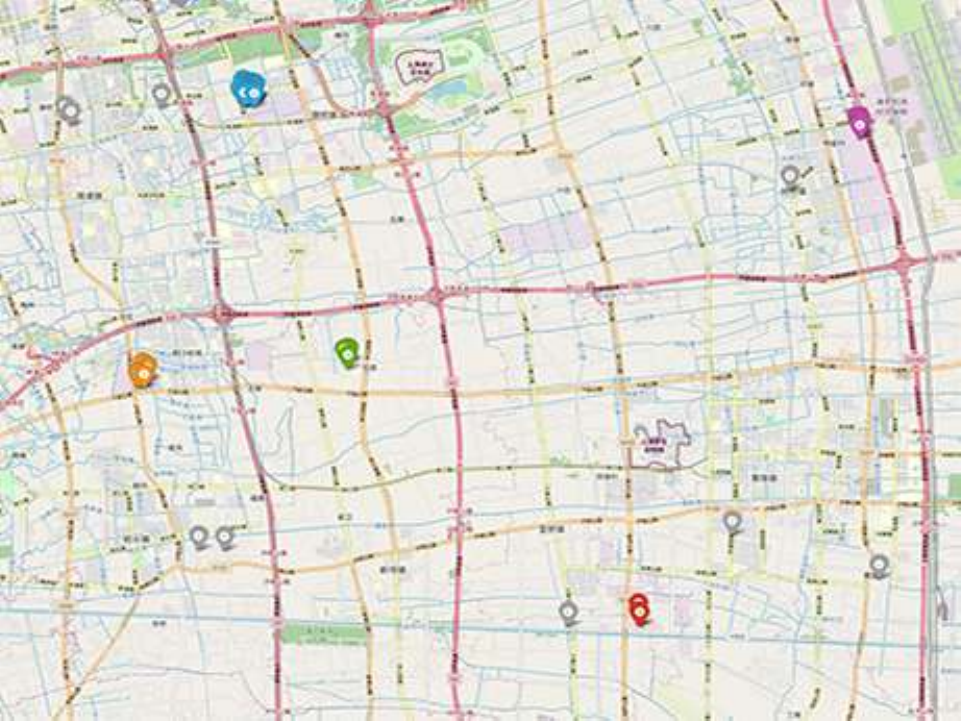}}
\caption{Number of clusters more than 1}
\label{fig:multiple_clusters}
\end{figure}

Among the 90,819 hosts in the \textit{clusters $\geq 2$} category, 82\% are successfully clustered. To gain deeper insights into the distribution of hosts in these batches, we compare the average intra-cluster distances with the minimum inter-cluster distances. The detailed results are shown in Fig.\ref{subfig:comparison}.

Observing the results, we note that the average intra-cluster distance within a batch is significantly smaller than the minimum inter-cluster distance. On average, the intra-cluster distance across all batches is 0.18 km, whereas the minimum inter-cluster distance is 2.5 km. This suggests that hosts under the same last-hop router are not evenly distributed geographically but are instead concentrated in specific areas, typically resembling the size of communities. Fig.\ref{subfig:multiple} shows a batch located in the Pudong District of Shanghai, where most hosts are grouped into five clusters. Within each cluster, the hosts are positioned in close proximity, often within the same community. In contrast, hosts from different clusters are separated by significant distances. A few remaining landmarks are not clustered due to insufficient density. Based on these observations, we can summarize the distribution characteristics of IP addresses: although most IP addresses are managed and allocated through the Dynamic Host Configuration Protocol (DHCP), their geographical distribution is not entirely random. Hosts under the same last-hop router tend to be grouped into "regions", which are often correlated with administrative boundaries, postal codes, or other geographic subdivisions.

\textbf{(1) Modeling IP Geolocation as a Classification Task:} We observe that hosts under the same last-hop router are often not uniformly distributed in geographical space but tend to cluster within certain regions, which aligns with the characteristics of classification problems. Treating IP geolocation as a classification task allows for greater focus on key regions, resulting in more accurate predictions of the region where an IP is located. Moreover, modeling IP geolocation as a classification task does not conflict with existing regression-based methods. Instead, it complements them by providing users with valuable information from a different perspective.

\textbf{(2) Considering Urban Spatial Division in IP Region Prediction:} Urban spatial division, including administrative regions, postal code areas, and census blocks, segments a city into regions of varying granularities based on specific standards. For example, Shanghai can be divided into "district-street-community" granularities according to administrative boundaries, with each granularity offering a unique perspective on regional organization. These regions may act as collective units in cyberspace. As noted in the previous analysis, many clusters are concentrated within geographical areas resembling community sizes, indicating a potential correlation between IP allocation and administrative boundaries. Incorporating urban spatial division into IP region prediction could significantly enhance accuracy. Additionally, urban spatial division itself provides valuable information that is often of interest to users.

\section{Proposed Methods} \label{Proposed Methods}
In this section, we provide a detailed introduction to the proposed HMCGeo framework. First, we define the key terms involved in this study, followed by a comprehensive explanation of the design and implementation of each component of the framework.
\subsection{Definition}
\textbf{Definition 1 (Multi-Granularity Geographical Region):}
The regions discussed in this paper refer to geographically defined areas based on specific criteria, which can be represented as Areas of Interest (AOIs) on a map. These criteria include administrative regions, postal codes, and census blocks. Based on different criteria, a city can be divided into regions of varying granularities. For example, according to administrative regions, Shanghai can be divided into 16, 228, and 5,998 regions at the district, street, and community granularities, respectively.

\textbf{Definition 2 (Landmark Hosts, Measurement Hosts, and Target Hosts):}
Landmark hosts are characterized by explicitly recorded IP and location information. However, they are typically passive, meaning they can be measured but cannot initiate measurements. Measurement hosts, in contrast, are globally distributed and equipped with active measurement capabilities, enabling them to send requests such as Ping and Traceroute to other hosts. Target hosts are those with known IP addresses but require their geographical regions to be determined.

\textbf{Definition 3 (IP Features):}
IP features consist of IP attribute features and IP measurement features. IP attribute features include properties retrieved from Whois queries, such as IPv4 address, ASN, ISP, and the associated country. Measurement features, on the other hand, comprise data obtained from Ping and Traceroute measurements conducted by measurement hosts, including metrics such as latency and hop count.

\textbf{Definition 4 (Region Prediction):}  
IP region prediction aims to determine the region where a target host is located by leveraging its IP features. Specifically, let there be $N$ landmarks, denoted as $L = \{ l_1, l_2, \dots, l_N \}$, with their corresponding IP features represented as $X_L$ and multi-granularity region labels as $\{ Y_L^1, Y_L^2, \dots, Y_L^G \}$, where $G$ denotes the number of granularities. The objective is to construct a data-driven model $F$ that, given the target IP features $X_T$, predicts the regions $\{ Y_T^1, Y_T^2, \dots, Y_T^G \}$ across all granularities. The entire inference process is defined as shown in Equation \ref{eq:region_prediction}.
\begin{equation} \label{eq:region_prediction}
    Y^1_T, Y^2_T, \dots, Y^G_T = \mathcal{F}(X_L, X_T, Y^1_L, Y^2_L, \dots, Y^G_L, \theta)
\end{equation}

Where $\theta$ represents the parameters of $F$, and in this paper, $F$ corresponds to HMCGeo. This method enables the prediction of the target host’s region across different granularities.

\subsection{HMCGeo}
\subsubsection{Overview of HMCGeo}
\begin{figure*}[!t]
\centering
\includegraphics[width=18cm]{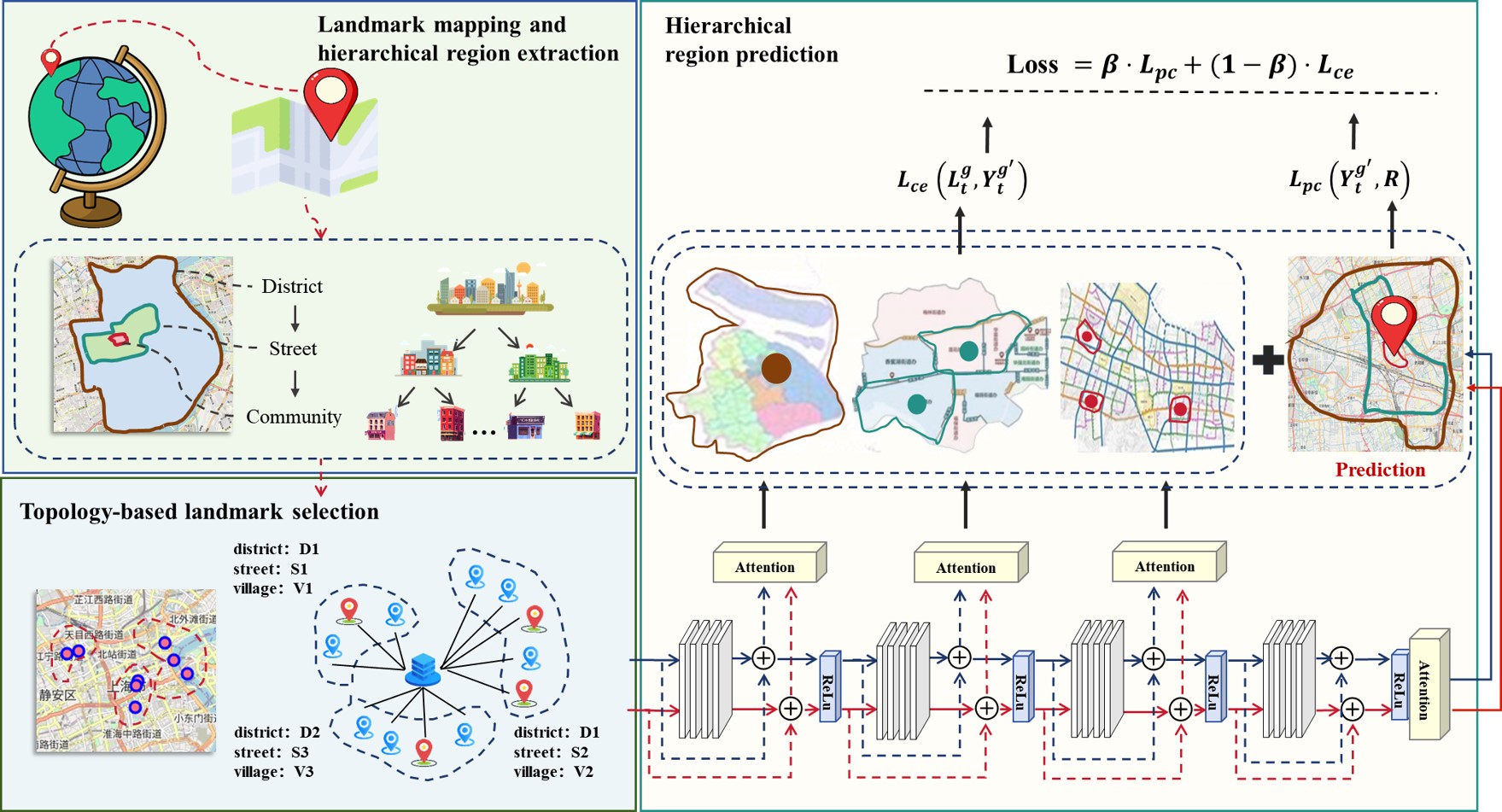}
\caption{The architecture of HMCGeo.}
\label{fig:HMCGeo}
\end{figure*}

As shown in Fig.\ref{fig:HMCGeo}, we propose a novel IP region prediction framework based on hierarchical multi-label classification, named HMCGeo. HMCGeo consists of three main components: \textbf{Landmark mapping and hierarchical region extraction}, \textbf{Topology-based landmark selection}, and \textbf{Hierarchical region prediction}. Landmark mapping assigns landmarks to regions at different granularities to generate labels, while hierarchical region extraction establishes relationships between regions across these granularities. The topology-based landmark selection leverages prior knowledge to select an appropriate set of landmarks for the target. Hierarchical region prediction employs residual connection-based feature extraction units to extract features and uses attention mechanisms to predict the region of the target host at each granularity. To balance the effectiveness of feature extraction at individual granularities and across the entire hierarchy, we adopt a fusion strategy combining local and global outputs. Additionally, we introduce probabilistic classification loss, which leverages the hierarchical relationships between regions at different granularities to further optimize prediction performance.

\subsubsection{Landmark Mapping and Hierarchical Region Extraction}
We obtain regional boundary information at various granularities for the target cities from public data portals and open-source platforms \cite{31}, \cite{32}, \cite{33}, \cite{34}. This information is typically represented as AOIs. We denote the boundaries as $\text{Boundary} = \{\text{Boundary}1, \text{Boundary}_2, \dots, \text{Boundary}_G\}$, where $G$ indicates the number of granularities, and $\text{Boundary}_g$ represents the set of regional boundaries at granularity $g$. For example, in Shanghai, there are three granularities: district, street, and community. In this case, $G=3$, and $\text{Boundary}_2$ corresponds to the AOI set of the 228 regions at the street granularity. Equation \ref{eq:location_mapping} demonstrates the process of mapping the coordinates of a target host, $\text{Coord}$, to regions at different granularities.
\begin{equation} \label{eq:location_mapping}
    Y^g_i = j \; \text{if} \; \text{Coord}_i \in \text{Boundary}^j_g
\end{equation}

Where $\text{coord}_l$ represents the coordinates of a landmark host. If these coordinates fall within the $j$-th AOI region, the landmark is assigned a region label of $j$ at granularity $i$. By comparing the coordinates with AOIs at different granularities, we derive the region labels of the landmark host for each granularity. Additionally, to represent the relationships between regions across granularities, we extract the hierarchical relationships $R$ as defined in Equation \ref{eq:hierarchical_relationship}.
\begin{equation} \label{eq:hierarchical_relationship}
    R_{ij} = 
    \begin{cases} 
    1 & \text{if region $i$ and $j$ overlap} \\
    0 & \text{otherwise}
    \end{cases}
\end{equation}

Where $i$ and $j$ represent two regions at different granularities. If there is an intersection between the two regions, they are considered to have a hierarchical relationship, and the corresponding position in $R$ is set to 1; otherwise, it is set to 0. The resulting $R$ is an $M \times M$ matrix, where $M$ is the total number of regions across all granularities, representing whether any two regions at different granularities have a hierarchical relationship.

\subsubsection{Topology-based Landmark Selection}
Landmarks play a critical role in IP geolocation, and selecting appropriate landmarks can significantly reduce prediction errors in fine-grained IP geolocation methods \cite{44}. GraphGeo proposes a rule-based IP host clustering method that constructs a graph from IP hosts sharing the same last-hop router for subsequent processing \cite{8}. Building on this approach, we perform topology-based landmark selection for the IP region prediction task, leveraging prior knowledge to identify a suitable set of landmark hosts for the target. Equation \ref{eq:clustering_condition} illustrates the process of selecting the landmark set $L_t$ for the target host $t$.
\begin{equation} \label{eq:clustering_condition}
    l \in L_t, \quad \text{if } l, t \in \text{same cluster}
\end{equation}

Where $l$ and $t$ denote a landmark host and a target host, respectively. The term $\text{cluster}$ refers to the results obtained from IP host clustering, in which all hosts within a cluster share the same last-hop visible router. Both $t$ and $L_t$ are input into subsequent processing, where an attention mechanism is used to model the relationship between the target host and the set of landmarks. This process facilitates the identification of the region associated with the target host $t$.

\subsubsection{Hierarchical Region Prediction}
Hierarchical region prediction determines the region of the target host at each granularity based on the set of landmark hosts. This process consists of hierarchical feature extraction and attention-based prediction. At each geographical granularity, the target's region is predicted using feature extraction units and attention mechanisms specific to that granularity. Hierarchical feature extraction processes the input features at each granularity using structurally similar extraction units, providing support for subsequent predictions. The feature extraction unit can adopt any type of architecture; for generality, we use a fully connected network in this study. Assume $X_l \in \mathbb{R}^{|D| \times N_l}$ and $X_t \in \mathbb{R}^{|D| \times N_t}$ represent the input features of the landmarks and target hosts, respectively, where $|D|$ denotes the feature dimension, and $N_l$ and $N_t$ denote the number of landmark and target hosts in the input. The feature extraction unit can then be expressed as shown in Equation 5.
\begin{equation}
H_l^1, H_t^1 = \sigma(W^1 (X_l, X_t) + b^1)
\label{eq:mlp1}
\end{equation}

Where $W_1$ and $b_1$ denote the weight and bias parameters, respectively. The activation function, $\sigma$, is set as Leaky ReLU in this study. $H_{l1}$ and $H_{t1}$ represent the outputs of the first feature extraction unit for landmark hosts and target hosts, respectively. To ensure effective information transfer across different granularities, we adopt residual connections, which help improve training stability and facilitate deeper learning \cite{29}. With the introduction of residual connections, the feature extraction unit can be formally expressed as shown in Equation 6.
\begin{equation}
H_l^g, H_t^g = \sigma(W^g (H_l^{(g-1)}, H_t^{(g-1)}) + b^g) + (H_l^{(g-1)}, H_t^{(g-1)})
\label{eq:mlp2}
\end{equation}

Where $g$ ranges from $1$ to $G+1$. The features extracted by the first $G$ feature extraction units are used to compute local outputs at each granularity, while the features from the $G+1$-th feature extraction unit are utilized to compute the global output. After feature extraction at each granularity, we employ an attention mechanism to predict the region of the target host at that specific granularity. Attention-based prediction captures the relationships between the target host and the landmark hosts at each granularity through attention units. These relationships are then used to predict the region of the target host at the current granularity. The attention unit is defined as shown in Equation 7.
\begin{equation}
Y_t^g = \text{Attention}^g (H_t^g, H_l^g, Y_l^g)
\label{eq:attention1}
\end{equation}

Where $Y_t^g$ represents the output vector at the $g$-th granularity, the region corresponding to the maximum value in the vector is considered the prediction result for that granularity. When $g$ falls within the range $[1, G]$, the length of $Y_t^g$ equals the number of regions at the $g$-th granularity. When $g$ equals $G+1$, the length of $Y_t^g$ equals the total number of regions across all granularities. $\text{attention}_g(Q, K, V)$ represents the attention mechanism used at the $g$-th layer \cite{30}, as defined in Equation 8.
\begin{equation}
\text{Attention}^g (Q, K, V) = \text{Softmax}\left(\frac{QK^T}{\sqrt{d_k}}\right)V
\label{eq:attention2}
\end{equation}

Where $Q = W_Q H_t^g$, $K = W_K H_l^g$, and $V = W_V Y_l^g$ represent the query, key, and value vectors, respectively. $d_k$ denotes the length of the key vector, serving as a normalization factor to adjust the sensitivity and learning effectiveness of the attention unit. $Q$, $K$, and $V$ correspond to the results of further feature extraction on the target features, landmark features, and landmark labels, respectively. The softmax-normalized result represents the similarity between the target and the landmark hosts.

To achieve more accurate region predictions, we adopt a strategy that combines local and global outputs. Specifically, the model predicts the local output (a single region) at each granularity, $Y_t^g, g \in [1, G]$, and additionally predicts the global output (multiple regions) at the finest granularity, represented as $Y_t^{G+1}$. The local prediction focuses on features specific to a single granularity, while the global prediction incorporates features across all granularities. To balance the contributions of local and global outputs in the final prediction, a hyperparameter $\alpha$ is introduced. This parameter adjusts the weight ratio between local and global outputs, enabling the model to account for single-granularity features and hierarchical relationships across multiple granularities. The final prediction result is defined as shown in Equation 9.

\begin{equation}
Y_t = \alpha \cdot \text{concat}(Y_t^1, Y_t^2, \ldots, Y_t^G) + (1 - \alpha) \cdot Y_t^{(G+1)}
\label{eq:output}
\end{equation}
\subsection{Loss Function}
We propose a composite loss function combining hierarchical cross-entropy loss and probabilistic classification loss to optimize model parameters during training. The hierarchical cross-entropy loss ensures that the predicted results are close to the ground truth labels, while the probabilistic classification loss leverages the hierarchical relationships between regions at different granularities to further refine the predictions. This approach not only improves the accuracy of region predictions but also enhances the model's ability to handle hierarchical multi-classification tasks. The hierarchical cross-entropy loss is defined as the sum of the cross-entropy losses at each granularity, as shown in Equation 10.

\begin{equation}
L_{ce} = \sum_{g=1}^G \xi(L_t^g, Y_t^{(g')})
\label{eq:ce1}
\end{equation}

Where $L_t^g$ represents the ground truth label at the $g$-th granularity, $Y_t^{g'}$ denotes the predicted result at the $g$-th granularity, and $\xi(\cdot)$ is the cross-entropy loss function, which is defined as shown in Equation 11.
\begin{equation}
\xi(p, q) = -\sum_{i=1}^n p(x_i) \log(q(x_i))
\label{eq:ce2}
\end{equation}

Where $n$ represents the total number of data samples in the batch. The divergence metric $p(x_i) \log (q(x_i))$ evaluates the deviation between the predicted results and the ground truth. By minimizing the cross-entropy loss, the model effectively learns the mapping relationships between the input features and the regions at different granularities.

The probabilistic classification loss improves the model prediction accuracy by leveraging hierarchical relationships among regions across different granularities to align predictions at each granularity \cite{30}. Assuming the total number of regions across all granularities is $M$, for a given input $x_t, X_l, Y_l$, the ground truth label is represented by $y$, which is an $M$-dimensional vector with values in $\{0,1\}$. The model output $y_{\text{pred}}$ is an $M$-dimensional vector corresponding to $y$. The unnormalized predicted probability $P(y_t \mid x_t, X_l, Y_l)$ is defined as shown in Equation 12.
\begin{equation}
P(y_t \mid x_t, X_l, Y_l) = \prod_{i=1}^N \phi_i (y_{pred}^i, y^i) \prod_{i,j \in \{1,2,\ldots,N\}} R_{i,j} (y_i, y_j)
\label{eq:pc1}
\end{equation}

Where $\phi_i(y_{\text{pred},i}, y_i) = e^{y_{\text{pred},i} \cdot \mathbb{I}(y_i = 1)}$ represents the evaluation function for the prediction result at the $i$-th region and the actual region. If $y_i = 1$, the value is $e^{y_{\text{pred},i}}$, otherwise it is 1. $R_{i,j}(y_i, y_j)$ denotes the constraint relationship between regions at different granularities. If region $j$ geographically contains region $i$, the value is 1; otherwise, it is 0. The probabilistic classification loss $L_{\text{pc}}$ is the boundary probability density that satisfies the hierarchical region constraint relationship, defined as $P(y_t = 1 \mid x_t, X_l, Y_l)/ Z(x_t, X_l, Y_l)$. Here, $Z(x_t, X_l, Y_l)$ is the normalization factor, used to normalize the joint probability density, as shown in Equation 13.

\begin{align}
Z(x_t, X_l, Y_l) = &\sum_{\bar{y} \in \{0,1\}^N} \left[ \prod_{i=1}^N \phi_i (y_{pred}^i, \bar{y}^i) \right. \nonumber\\
&\left. \prod_{i,j \in \{1,2,\ldots,N\}} R_{i,j} (\bar{y}_i, \bar{y}_j) \right]
\label{eq:pc2}
\end{align}


Where $\bar{y}$ represents the set of all region results that satisfy the hierarchical constraints. The final loss function, $\text{Loss}$, is weighted and balanced by the hyperparameter $\beta$ to combine the hierarchical cross-entropy loss $L_{\text{ce}}$ and the probabilistic classification loss $L_{\text{pc}}$, as shown in Equation 14.
\begin{equation}
Loss = \beta \cdot L_{pc} + (1 - \beta) \cdot L_{ce}
\label{eq:loss}
\end{equation}

Where the hyperparameter $\beta$ is responsible for weighting and balancing the two types of losses. When $\beta$ is small, the model focuses more on the classification accuracy at each granularity. Conversely, when $\beta$ is larger, the model places more emphasis on the hierarchical constraints between different granularities. In Section \ref{Sensitivity Analysis}, we evaluate the region prediction performance under different values of $\beta$.
\section{Experiments} \label{Experiments}
In this section, we carefully design a series of experiments to evaluate the performance of HMCGeo, including region prediction performance analysis, ablation experiments, case analysis, geolocation error analysis, and sensitivity analysis. We first provide an overview of the experimental setup, followed by an explanation of the design and results for each experiment.
\subsection{Experimental Setup}
\subsubsection{Datasets}
We conduct experiments on real-world public IP geolocation datasets from three cities: New York, Los Angeles, and Shanghai. TABLE \ref{tab:datasets} presents the basic details of these three datasets.

\begin{table}[h]
  \centering
  \caption{Datasets Statistics}
  \label{tab:datasets}
  \setlength{\tabcolsep}{5pt}
  \renewcommand{\arraystretch}{1.2} 
  \begin{tabular}{cccccc}
    \hline
    Datasets & IPs & Dimensions & Batchs & Hierarchy & Class \\
    \hline
    \multirow{4}{*}{New York} & \multirow{4}{*}{91809} & \multirow{4}{*}{31} & \multirow{4}{*}{1872} & Boro & 5\\
                        &                        &                     &                      & CDTA & 70\\
                        &                        &                     &                      & NTA  & 253\\
                        &                        &                     &                      & CT   & 2240\\
    \hline
    \multirow{4}{*}{Los Angeles} & \multirow{4}{*}{92803} & \multirow{4}{*}{31} & \multirow{4}{*}{2240} & City & 69\\
                        &                        &                     &                       & ZIP  & 241\\
                        &                        &                     &                       & CT   & 1753\\
                        &                        &                     &                       & BG   & 3676\\
    \hline
    \multirow{3}{*}{Shanghai} & \multirow{3}{*}{126258} & \multirow{3}{*}{52} & \multirow{3}{*}{8818} & District & 16\\
                        &                        &                      &                       & Street  & 228\\
                        &                        &                      &                       & Community & 5998\\
    \hline
  \end{tabular}
\end{table}

\begin{table*}
  \centering
  \caption{F1 metrics(Precision/Recall/F1 score)/Accuracy Results on Datasets by Comparing to SOTA Methods.}
  \label{hmcgeo_performance}
  \setlength{\tabcolsep}{3.5pt}
  \renewcommand{\arraystretch}{1.15} 
  \begin{tabular}{cccccccccccccc}
  \hline
    \multirow{2}{*}{Datasets} & \multirow{2}{*}{Hierarchy} & \multicolumn{10}{c}{Macro-Averaged Precision/Recall/F1 Score and Accuracy} \\
        
    \cline{3-14}

    &  & \multicolumn{2}{c}{GraphGeo} & \multicolumn{2}{c}{RIPGeo} & \multicolumn{2}{c}{TrustGeo} & \multicolumn{2}{c}{HMCGeo} & \multicolumn{2}{c}{HMCGeo-2} & \multicolumn{2}{c}{HMCGeo-3} \\
    
    \hline
    \multirow{12}{*}{New York} & \multirow{3}{*}{Boro} & 0.965 & \multirow{3}{*}{0.967} & 0.973 & \multirow{3}{*}{\textbf{0.973}} & 0.973 & \multirow{3}{*}{0.972} & 0.973 & \multirow{3}{*}{0.973} & 0.581 & \multirow{3}{*}{\textbf{0.993}} & 0.444 & \multirow{3}{*}{\textbf{0.997}}\\
                         &  & 0.972 &  & \textbf{0.977} &  & 0.975 &  & 0.976 &  & \textbf{0.992} &  & \textbf{0.995} &\\
                         &  & 0.968 &  & \textbf{0.975} &  & 0.974 &  & 0.975 &  & 0.715 &  & 0.575 &\\
        \cline{2-14}
     & \multirow{3}{*}{CDTA} & 0.501 & \multirow{3}{*}{0.654} & 0.498 & \multirow{3}{*}{0.665} & 0.531 & \multirow{3}{*}{0.685} & 0.536 & \multirow{3}{*}{\textbf{0.729}} & 0.402 & \multirow{3}{*}{\textbf{0.894}} & 0.307 & \multirow{3}{*}{\textbf{0.947}}\\
                         &  & 0.490 &  & 0.502 &  & 0.505 &  & \textbf{0.516} &  & \textbf{0.726} &  & \textbf{0.803} &\\
                         &  & 0.468 &  & 0.478 &  & 0.482 &  & \textbf{0.506} &  & 0.486 &  & 0.412 &\\
        \cline{2-14}
      & \multirow{3}{*}{NTA} & 0.200 & \multirow{3}{*}{0.403} & 0.219 & \multirow{3}{*}{0.417} & 0.231 & \multirow{3}{*}{0.436} & 0.203 & \multirow{3}{*}{\textbf{0.531}} & 0.201 & \multirow{3}{*}{\textbf{0.738}} & 0.181 & \multirow{3}{*}{\textbf{0.830}}\\
                         &  & 0.227 &  & 0.233 &  & \textbf{0.235} &  & 0.234 &  & \textbf{0.403} &  & 0.515 &\\
                         &  & 0.175 &  & 0.188 &  & 0.194 &  & \textbf{0.201} &  & \textbf{0.252} &  & \textbf{0.252} &\\
        \cline{2-14}
     & \multirow{3}{*}{CT} & 0.019 & \multirow{3}{*}{0.095} & 0.021 & \multirow{3}{*}{0.096} & 0.025 & \multirow{3}{*}{0.125} & 0.021 & \multirow{3}{*}{\textbf{0.373}} & 0.022 & \multirow{3}{*}{0\textbf{.491}} & 0.023 & \multirow{3}{*}{\textbf{0.540}}\\
                         &  & 0.033 &  & 0.033 &  & 0.032 &  & \textbf{0.033} &  & \textbf{0.060} &  & \textbf{0.086} &\\
                         &  & 0.014 &  & 0.015 &  & 0.018 &  & \textbf{0.023} &  & \textbf{0.028} &  & \textbf{0.031} &\\
                         
    \hline
    \multirow{12}{*}{Los Angeles} & \multirow{3}{*}{City} & 0.244 & \multirow{3}{*}{0.858} & 0.251 & \multirow{3}{*}{0.876} & 0.257 & \multirow{3}{*}{0.884} & 0.190 & \multirow{3}{*}{\textbf{0.892}} & 0.155 & \multirow{3}{*}{\textbf{0.967}} & 0.160 & \multirow{3}{*}{\textbf{0.980}}\\
                         &  & 0.206 &  & 0.212 &  & \textbf{0.227} &  & 0.174 &  & \textbf{0.235} &  & \textbf{0.297} &\\
                         &  & 0.195 &  & 0.213 &  & \textbf{0.222} &  & 0.171 &  & 0.156 &  & 0.158 &\\                     
        \cline{2-14}
     & \multirow{3}{*}{ZIP} & 0.195 & \multirow{3}{*}{0.423} & 0.212 & \multirow{3}{*}{0.432} & 0.212 & \multirow{3}{*}{0.444} & 0.210 & \multirow{3}{*}{\textbf{0.507}} & 0.151 & \multirow{3}{*}{\textbf{0.679}} & 0.135 & \multirow{3}{*}{\textbf{0.785}}\\
                         &  & 0.170 &  & 0.195 &  & 0.197 &  & \textbf{0.206} &  & \textbf{0.240} &  & \textbf{0.292} &\\
                         &  & 0.158 &  & 0.178 &  & 0.181 &  & \textbf{0.192} &  & 0.170 &  & 0.165 &\\
        \cline{2-14}
      & \multirow{3}{*}{CT} & 0.030 & \multirow{3}{*}{0.094} & 0.036 & \multirow{3}{*}{0.104} & 0.037 & \multirow{3}{*}{0.080} & 0.033 & \multirow{3}{*}{\textbf{0.353}} & 0.020 & \multirow{3}{*}{\textbf{0.464}} & 0.019 & \multirow{3}{*}{\textbf{0.523}}\\
                         &  & 0.034 &  & 0.039 &  & 0.035 &  & \textbf{0.045} &  & 0.044 &  & \textbf{0.056} &\\
                         &  & 0.021 &  & 0.025 &  & 0.023 &  & \textbf{0.034} &  & 0.024 &  & 0.024 &\\
        \cline{2-14}
     & \multirow{3}{*}{BG} & 0.015 & \multirow{3}{*}{0.052} & 0.017 & \multirow{3}{*}{0.047} & 0.016 & \multirow{3}{*}{0.038} & 0.016 & \multirow{3}{*}{\textbf{0.336}} & 0.009 & \multirow{3}{*}{\textbf{0.439}} & 0.008 & \multirow{3}{*}{\textbf{0.496}}\\
                         &  & 0.015 &  & 0.018 &  & 0.017 &  & \textbf{0.024} &  & 0.021 &  & \textbf{0.027} &\\
                         &  & 0.009 &  & 0.011 &  & 0.010 &  & \textbf{0.018} &  & 0.011 &  & 0.011 &\\
                         
    \hline
    \multirow{9}{*}{Shanghai} & \multirow{3}{*}{District} & 0.699 & \multirow{3}{*}{0.806} & 0.772 & \multirow{3}{*}{0.848} & 0.666 & \multirow{3}{*}{0.823} & 0.804 & \multirow{3}{*}{\textbf{0.866}} & 0.461 & \multirow{3}{*}{\textbf{0.925}} & 0.364 & \multirow{3}{*}{\textbf{0.946}}\\
                         &  & 0.672 &  & 0.758 &  & 0.636 &  & \textbf{0.776} &  & \textbf{0.875} &  & 0\textbf{.912} &\\
                         &  & 0.680 &  & 0.759 &  & 0.645 &  & \textbf{0.787} &  & 0.568 &  & 0.469 &\\
        \cline{2-14}
     & \multirow{3}{*}{Street} & 0.307 & \multirow{3}{*}{0.311} & 0.296 & \multirow{3}{*}{0.333} & 0.210 & \multirow{3}{*}{0.269} & 0.321 & \multirow{3}{*}{\textbf{0.416}} & 0.237 & \multirow{3}{*}{\textbf{0.572}} & 0.186 & \multirow{3}{*}{\textbf{0.665}}\\
                         &  & 0.195 &  & 0.210 &  & 0.148 &  & \textbf{0.245} &  & \textbf{0.366} &  & \textbf{0.445} &\\
                         &  & 0.190 &  & 0.201 &  & 0.139 &  & \textbf{0.241} &  & \textbf{0.260} &  & 0.241 &\\
        \cline{2-14}
      & \multirow{3}{*}{Community} & 0.048 & \multirow{3}{*}{0.129} & 0.036 & \multirow{3}{*}{0.136} & 0.008 & \multirow{3}{*}{0.015} & 0.029 & \multirow{3}{*}{\textbf{0.189}} & 0.017 & \multirow{3}{*}{\textbf{0.233}} & 0.014 & \multirow{3}{*}{\textbf{0.261}}\\
                         &  & 0.034 &  & 0.027 &  & 0.007 &  & \textbf{0.035} &  & \textbf{0.035} &  & \textbf{0.043} &\\
                         &  & \textbf{0.031} &  & 0.025 &  & 0.005 &  & 0.026 &  & 0.018 &  & 0.018 &\\
    \hline
  \end{tabular}
\end{table*}

The datasets for New York, Los Angeles, and Shanghai consist of 91,809, 92,803, and 126,258 IP addresses, respectively. We randomly select 80\% of the data for training and use the remaining 20\% for testing, ensuring consistent data partitioning in the comparative experiments. The feature dimensions for the New York and Los Angeles datasets are 31, while the Shanghai dataset has a feature dimension of 52. Using the IP host aggregation method, we group all hosts into 1,872, 2,240, and 8,818 clusters, respectively, and assign landmark host sets to the target hosts within each cluster based on the topology-based landmark selection method. Additionally, we segment each city into multiple granularities based on administrative divisions, postal codes, etc. Specifically, in decreasing order of region size, New York is divided into Borough (Boro), Community District Tabulation Area (CDTA), Neighborhood Tabulation Area (NTA), and Census Tract (CT) granularities, containing 5, 70, 253, and 2,240 regions, respectively. Los Angeles is divided into City Boundaries (City), Zip Code Boundaries (ZIP), Census Tract (CT), and Census Block Groups (BG) granularities, containing 69, 241, 1,753, and 3,676 regions, respectively. Shanghai is divided into District, Street, and Community granularities, with 16, 228, and 5,998 regions, respectively \cite{31}, \cite{32}, \cite{33}, \cite{34}. This study is the first to focus on IP geolocation at the community granularity.
\subsubsection{Implementation Details}
HMCGeo is built on Pytorch 1.8. The training and inference platform uses an Intel(R) Core(TM) i9-10980XE CPU @ 3.00GHz as the processor, an NVIDIA GeForce RTX 3090 (24GB) as the graphics processing unit, and runs on the Ubuntu 22.04 operating system. The model optimization is handled by the Adam optimizer, and the model parameters are initialized using Pytorch's default initializer. To accurately configure the model hyperparameters, we performed the following steps: we used grid search to determine the optimal encoding feature hidden layer length from the candidate set $[16, 32, 48, 64, 128]$; we searched for the best learning rate from the parameter set $[2\text{e-}2, 1\text{e-}2, 2\text{e-}3, 1\text{e-}3, 2\text{e-}4, 1\text{e-}4]$; we explored the optimal values of $\alpha$ and $\beta$ within the range $[0, 0.05, 0.1, \dots, 0.95, 1]$; and we determined the cross-entropy loss ratio coefficient for each granularity within the range $[0.1, 0.5, 1.0]$.

\begin{table*}[h]
  \centering
  \caption{Accuracy Results of Ablation Study}
  \label{ablation_study}
  \setlength{\tabcolsep}{3.5pt}
  \renewcommand{\arraystretch}{1.15}
  \begin{tabularx}{\textwidth}{>{\centering\arraybackslash}X >{\centering\arraybackslash}X >{\centering\arraybackslash}X >{\centering\arraybackslash}X >{\centering\arraybackslash}X >{\centering\arraybackslash}X >{\centering\arraybackslash}X}
    \hline
        \multirow{2}{*}{Datasets} & \multirow{2}{*}{Hierarchy} & \multicolumn{5}{c}{Accuracy} \\
        
        \cline{3-7}
        
         &  & w/o Triplets & w/o Dynamic & w/o GCN & w/o MLP & Ours \\
        
        \hline
        \multirow{4}{*}{New York} & Boro & 0.9756 & 0.9755 & 0.9757 & 0.9741 & \textbf{0.9758}\\
                            & CDTA & 0.7330 & 0.7335 & 0.7352 & 0.7287 & \textbf{0.7370}\\
                            & NTA & 0.5300 & 0.5304 & 0.5360 & 0.5238 & \textbf{0.5386}\\
                            & CT & 0.3824 & 0.3753 & 0.3761 & 0.3679 & \textbf{0.3831}\\
        \hline    
        \multirow{4}{*}{Los Angeles} & City & 0.9063 & 0.9074 & 0.9076 & 0.9043 & \textbf{0.9081}\\
                            & ZIP & 0.5322 & 0.5338 & 0.5339 & 0.5185 & \textbf{0.5370}\\
                            & CT & 0.3726 & 0.3695 & 0.3765 & 0.3626 & \textbf{0.3792}\\
                            & BG & 0.3598 & 0.3576 & 0.3656 & 0.3493 & \textbf{0.3690}\\
        \hline
        \multirow{3}{*}{Shanghai} & District & 0.8800 & 0.8824 & 0.8845 & 0.8706 & \textbf{0.8850}\\
                            & Street & 0.4331 & 0.4332 & 0.4364 & 0.4200 & \textbf{0.4388}\\
                            & Community & 0.2139 & 0.2134 & 0.2160 & 0.2060 & \textbf{0.2164}\\
        \hline
  \end{tabularx}
\label{fig:case0}
\end{table*}

\subsubsection{Baselines}
To evaluate the performance of HMCGeo in IP geolocation tasks, we conduct a systematic comparison with current state-of-the-art IP geolocation methods, including GraphGeo, RIPGeo, and TrustGeo.

GraphGeo \cite{8} uses GNNs for street-level IP geolocation. This method builds a weighted graph using network topology information, connecting various IP hosts, enabling the target host to aggregate knowledge from neighboring nodes. Additionally, GraphGeo employs an uncertainty-aware GNN mechanism, allowing for more accurate and stable geolocation predictions in complex network environments.

RIPGeo \cite{20} focuses on prediction robustness and designs a reliable framework for street-level IP geolocation. This framework uses self-supervised adversarial training strategies to simulate disturbances in network measurements, enhancing the stability and robustness of IP geolocation. Furthermore, RIPGeo incorporates a multi-task learning framework to address the homogenization issue caused by self-supervised training, improving both the positioning performance and the system's ability to resist interference.

TrustGeo \cite{19}, on the other hand, emphasizes uncertainty evaluation in prediction results. The method integrates deep evidential learning and graph neural networks to achieve accurate and reliable predictions. It introduces dynamic graph learning and multi-perspective information fusion, which better integrates the relationships between the target host and its neighboring hosts, thus optimizing geolocation performance. Additionally, TrustGeo introduces confidence scores to assess the reliability of the predictions.

\subsubsection{Evaluation Metrics}
This paper primarily focuses on the performance of predicting the region of the target host, treating it as a classification task. Therefore, the evaluation of region prediction results is based on accuracy, Macro-Averaged Precision, Recall, and F1 Score, as shown in Equations 14–16. Accuracy refers to the proportion of correctly predicted samples out of the total samples. Macro-Averaged Precision and Recall represent the model's ability to correctly identify positive samples and recognize all positive samples, respectively. The Macro-Averaged F1 score is a comprehensive evaluation metric that balances precision and recall, providing an overall assessment of the model performance. Additionally, in Section \ref{Geolocation Error Analysis}, we focus on the CDF performance comparison across different methods.
\begin{equation}
\text{precision} = \frac{1}{N} \sum_{i=1}^N \frac{TP_i}{TP_i + FP_i}
\label{eq:precision}
\end{equation}
\begin{equation}
\text{recall} = \frac{1}{N} \sum_{i=1}^N \frac{TP_i}{TP_i + FN_i}
\label{eq:recall}
\end{equation}
\begin{equation}
\text{F1 score} = \frac{1}{N} \sum_{i=1}^N \frac{2 \cdot (\text{recall}_i \cdot \text{precision}_i)}{\text{recall}_i + \text{precision}_i}
\label{eq:f1}
\end{equation}

\subsection{Analysis of IP Geolocation Performance} \label{Analysis of IP Geolocation Performance}
To evaluate the region prediction performance of \textit{HMCGeo}, we compare it with the current state-of-the-art methods: \textit{GraphGeo}, \textit{RIPGeo}, and \textit{TrustGeo}. Specifically, we focus on the accuracy and F1 metrics for region prediction at each granularity, as shown in TABLE \ref{hmcgeo_performance}. We map the coordinates output by the current methods to regions at different granularities for comparison. For example, if the predicted coordinates are $[121.4274, 31.1318]$, which correspond to the Xuhui District, Lingyun Road Street, and Xingrongyuan Community in Shanghai, we use \textit{``Xuhui District''}, \textit{``Lingyun Road Street''}, \textit{``Xingrongyuan Community''} as the corresponding region prediction result. Additionally, to demonstrate the potential of \textit{HMCGeo}, we evaluate the prediction performance by selecting multiple regions at each granularity. If the correct region is included in the predicted regions, it is considered a correct prediction. \textit{HMCGeo-2} and \textit{HMCGeo-3} represent the performance when selecting 2 and 3 regions, respectively.

HMCGeo achieves the best performance across all three datasets. The experiments reveal the following: compared to other methods, HMCGeo shows significant improvements in nearly all granularities, with the improvements becoming more pronounced as granularity increases. At coarser granularities (City and District), HMCGeo's region prediction accuracy improves by 1\%-2\% compared to the next best method, while at finer granularities (CT, BG, and Community), the improvement is substantial, ranging from 47.6\% to 546.7\%. In terms of F1 score, HMCGeo achieves the highest recall and F1 scores across most granularities. Except for slight gaps at the Boro, NTA, and City granularities, HMCGeo outperforms the next best method in other granularities with average improvements of 8.8\% in recall and 14.6\% in F1 score. HMCGeo-2 and HMCGeo-3 show improvements in accuracy and recall but a decline in precision.

\begin{figure*}[h]\label{table:11}
    \centering
    \subfloat[\scriptsize\label{subfig:a}]{\includegraphics[width=0.32\textwidth]{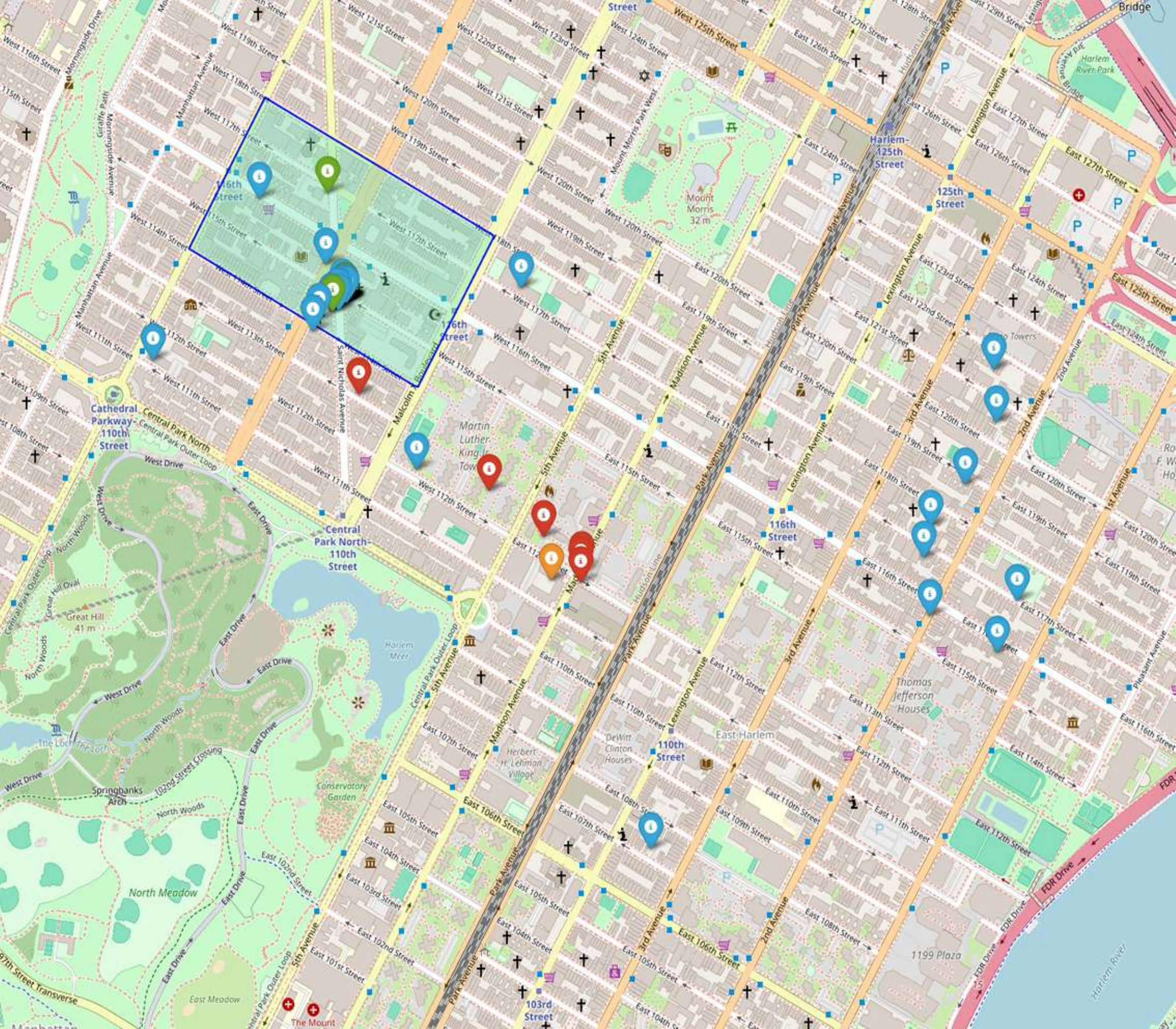}}\hfill
    \subfloat[\scriptsize\label{subfig:b}]{\includegraphics[width=0.32\textwidth]{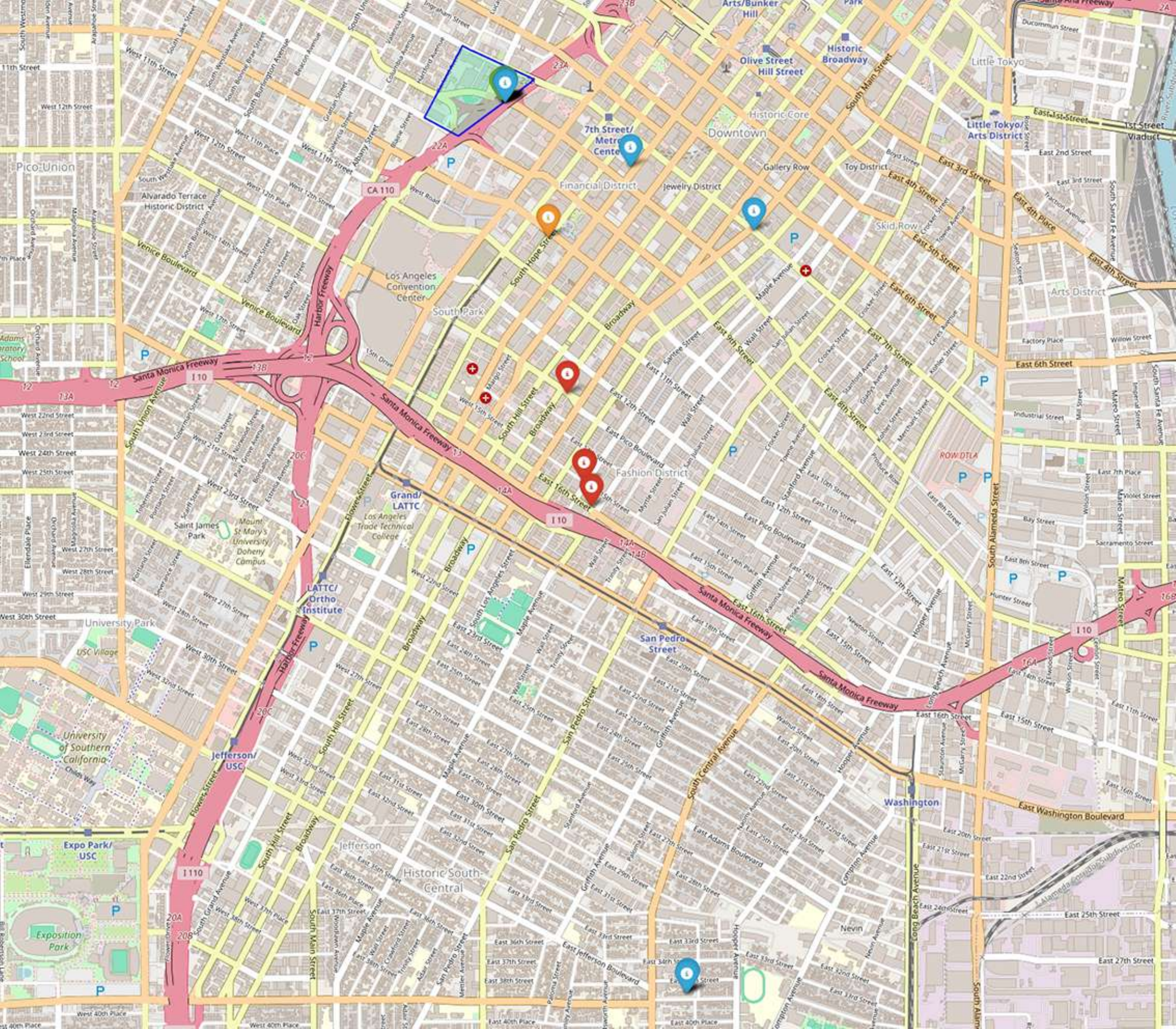}}\hfill
    \subfloat[\scriptsize\label{subfig:c}]{\includegraphics[width=0.32\textwidth]{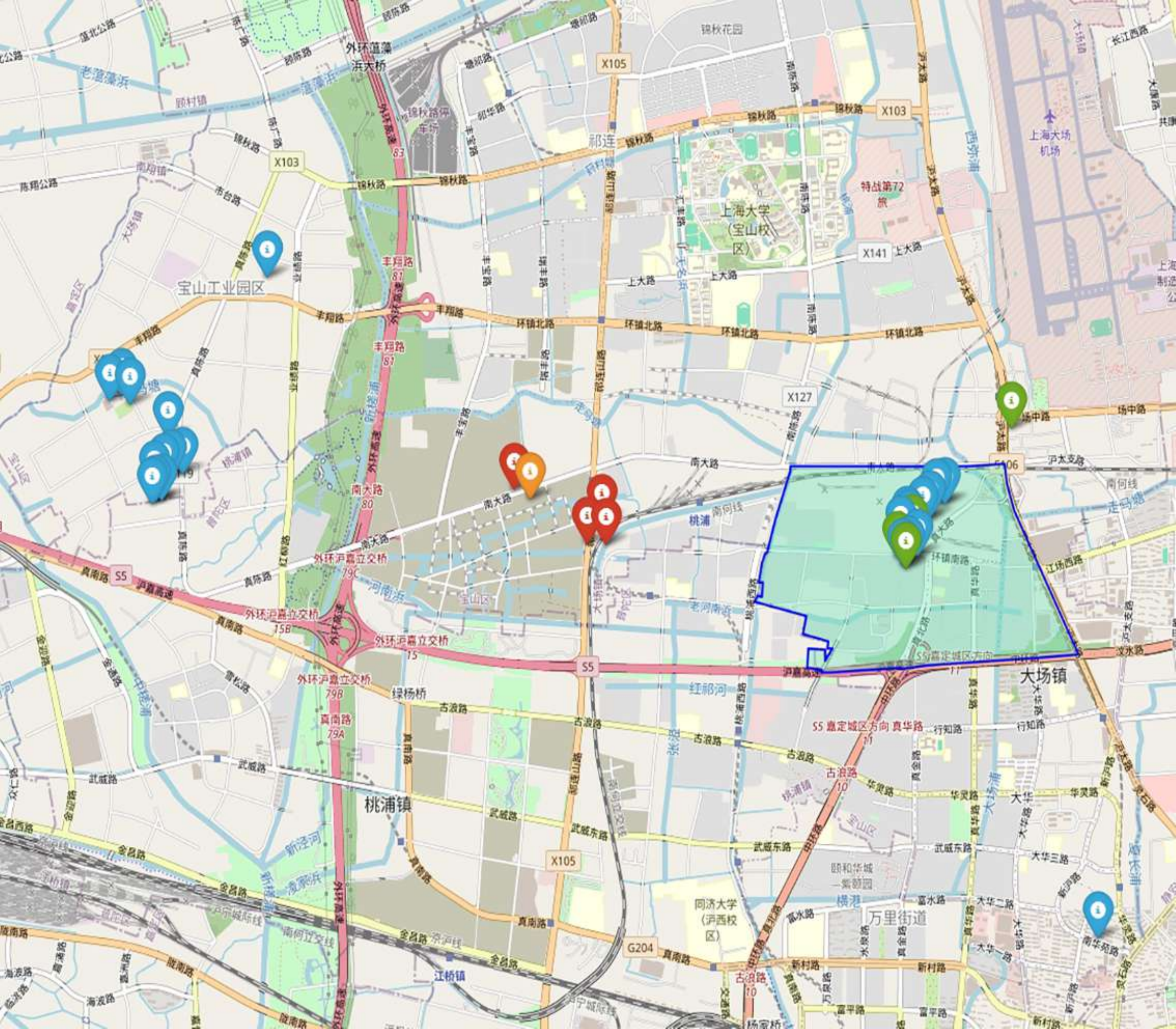}}\\
    \subfloat[\scriptsize\label{subfig:d}]{\includegraphics[width=0.32\textwidth]{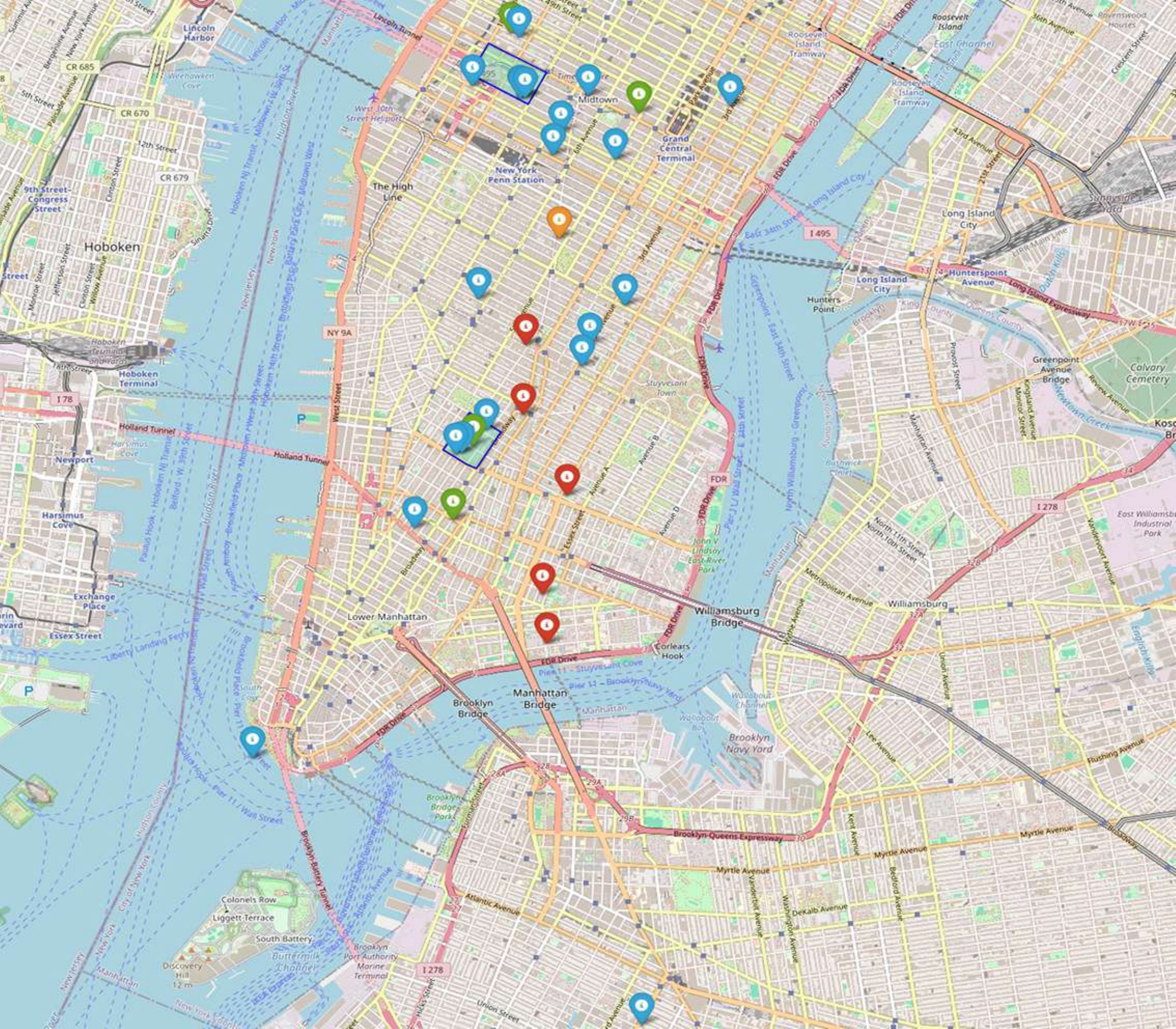}}\hfill
    \subfloat[\scriptsize\label{subfig:e}]{\includegraphics[width=0.32\textwidth]{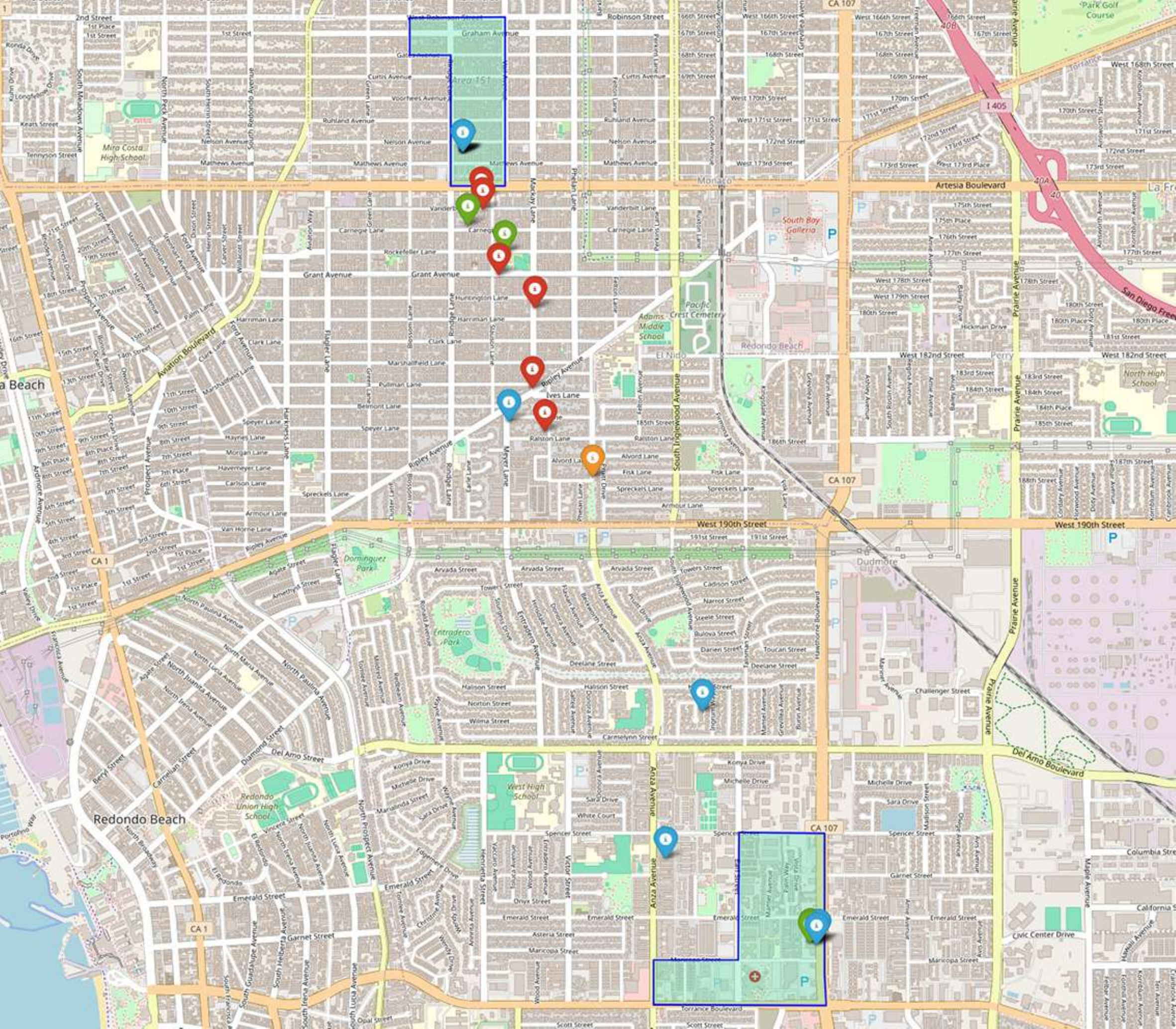}}\hfill
    \subfloat[\scriptsize\label{subfig:f}]{\includegraphics[width=0.32\textwidth]{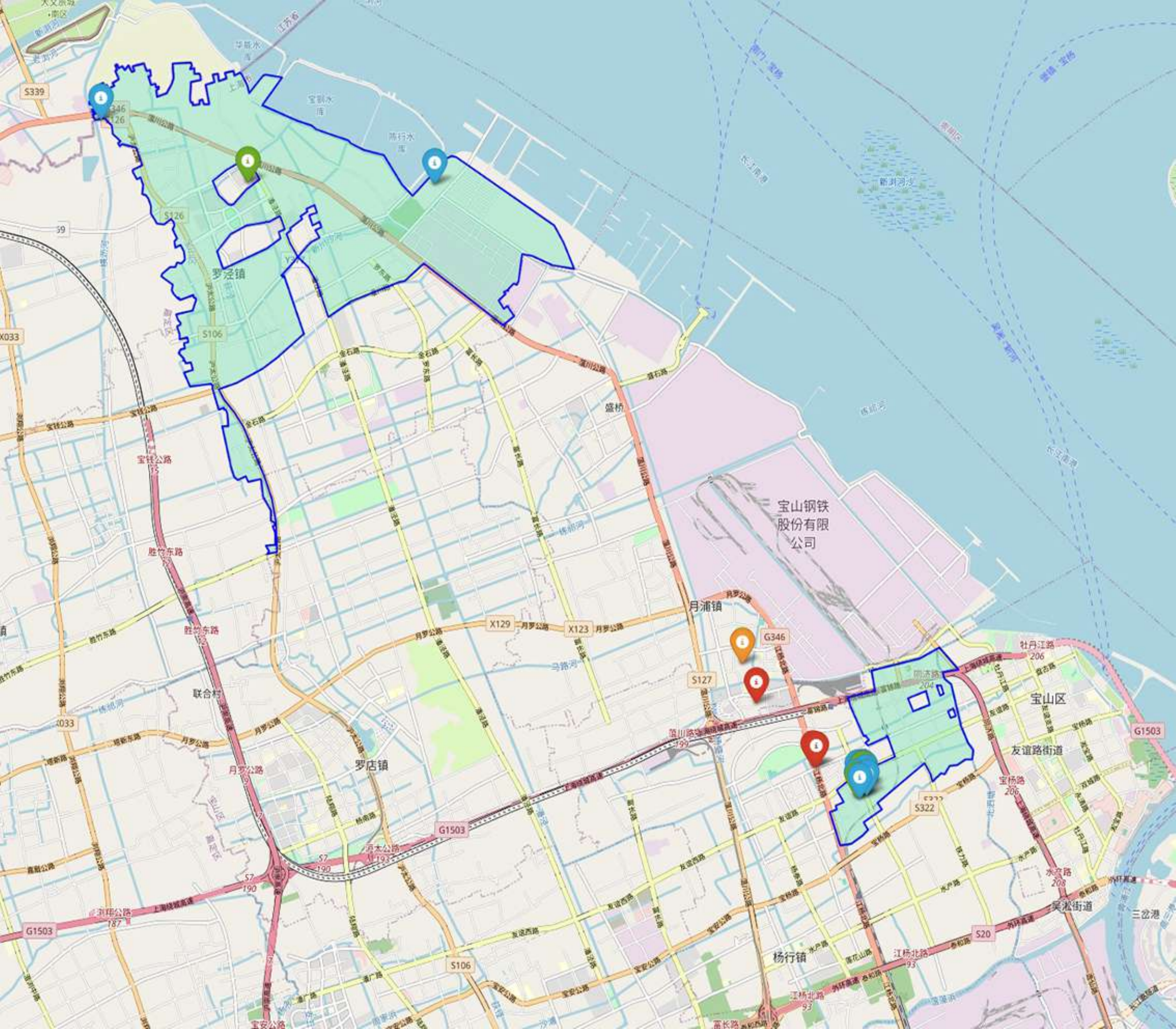}}
    \caption{Comparison of geolocation performance between HMCGeo and RIPGeo, where figures a, b, c and figures d, e, f represent the geolocation performance on the New York, Los Angeles, and Shanghai datasets for target aggregation and target dispersion, respectively.}
    \label{case}
\end{figure*}

We attribute these results to the following factors: GraphGeo, RIPGeo, and TrustGeo approach IP geolocation from different angles, such as model structure, robustness, and uncertainty. However, they treat the problem as a regression task and optimize with MSE loss. Previous studies have shown that IP features contain noise due to unreliable attribute sources and noisy measurement environments, which prevents neural networks from accurately learning the mapping between IP features and coordinates, leading to predictions biased toward the midpoint of landmark host locations. HMCGeo, by treating IP geolocation as a region prediction task, avoids this issue and delivers stronger region prediction performance. As granularity becomes finer, the number of regions increases and their size decreases, amplifying the limitations of existing methods, which tend to predict near the midpoint of landmark locations. With more "opportunities" to predict the correct region at different granularities, HMCGeo increases true positives (TP), resulting in a significant improvement in recall. However, due to IP feature noise, HMCGeo may still produce incorrect predictions, affecting precision and preventing substantial improvement in precision compared to other methods. Nonetheless, in the F1 score metric, our method demonstrates superior region prediction capability and more accurate identification of the target host's region. HMCGeo-2 and HMCGeo-3 show higher recall, but their precision declines because they predict multiple regions as potential results, increasing both TP and false positives (FP).

\subsection{Ablation Experiments}
To evaluate the contributions of the composite loss and output fusion strategy in HMCGeo, we conduct ablation experiments in two parts. First, we remove CE Loss and PC Loss separately while keeping other components unchanged to assess their impact on performance. Then, we evaluate the effect of different output strategies by retaining only the local and global outputs. The results are shown in TABLE \ref{ablation_study}.

By comparing the results of w/o CE Loss and w/o PC Loss, we observe that, in most cases, the region prediction performance follows this order: HMCGeo > w/o PC Loss > w/o CE Loss. Only at the coarse granularities (CDTA, City, District) across the three datasets does w/o PC Loss outperform. The hierarchical cross-entropy loss focuses on optimizing the model to predict regions close to the correct ones, while probabilistic classification loss also ensures that predictions follow hierarchical relationships across granularities. This makes the performance of cross-entropy loss better when used alone. In contrast, the composite loss balances both losses, improving predictions through hierarchical constraints while maintaining accuracy, with notable improvements in the Los Angeles dataset (7.7\%, 8.4\%, and 6.4\% higher accuracy at the ZIP, CT, and BG granularities, respectively). Regarding the performance of w/o Global and w/o Local, we observe that, in most granularities, the region prediction order is: HMCGeo > w/o Local > w/o Global. Only at the Boro, CT (LA), and BG granularities does w/o Global perform better. The residual connection-based feature extraction units pass coarse-grained features to finer granularities. Local output relies on features from the current granularity, while global output uses features from the finest granularity, often leading to better performance. The output fusion strategy combines both outputs, integrating the features of each granularity and further optimizing prediction results.

    


\begin{figure*}[h]
\centering
\includegraphics[width=18cm]{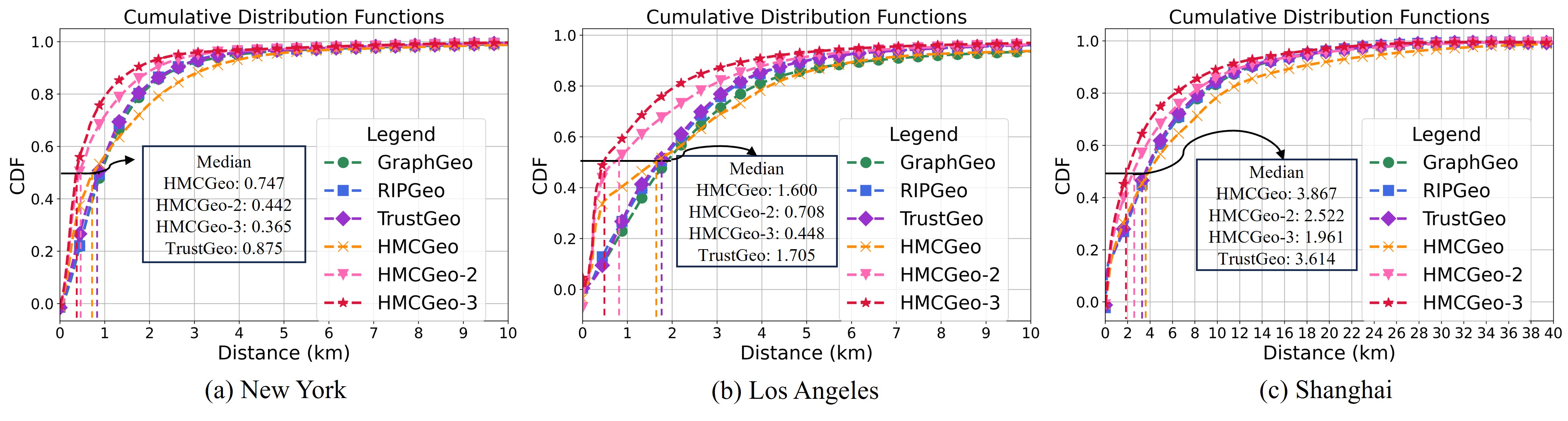}
\caption{Comparison of Geolocation Errors.}
\label{fig:Geolocation Error}
\end{figure*}

\subsection{Case Analysis} \label{Case Analysis}
To visually compare the region prediction performance of HMCGeo, we design a case analysis experiment. This experiment visualizes the prediction results of both HMCGeo and RIPGeo to contrast the outcomes of classification and regression methods. The results are shown in Fig.\ref{case}, where Fig.\ref{subfig:a}, Fig.\ref{subfig:b}, and Fig.\ref{subfig:c} depict cases where the target host is located in a single region, and Fig.\ref{subfig:d}, Fig.\ref{subfig:e}, and Fig.\ref{subfig:f} show cases where the target host is located across multiple regions.

The prediction results of RIPGeo are "points", while HMCGeo predicts "regions". From these cases, we observe the following:.

\begin{itemize}
    \item {RIPGeo} predictions are often close to the landmark center, while HMCGeo predictions are unaffected by it. As shown in Fig.\ref{subfig:a} and Fig.\ref{subfig:c}, most red coordinates cluster around the orange points, even without information about the corresponding region. In contrast, the green regions are unaffected by the orange coordinates, leading to accurate predictions of the target host's region.

    \item {RIPGeo} predictions are more sensitive to outlier landmarks, causing biased results. HMCGeo is less influenced by outliers. Fig.\ref{subfig:b} and Fig.\ref{subfig:d} show that RIPGeo's prediction is swayed by an outlier landmark located at the bottom of the image, shifting the prediction from the landmark center. However, HMCGeo predictions remain unaffected by this outlier.

    \item {RIPGeo} struggles in regions with few landmarks, while \textit{HMCGeo} still makes accurate predictions with sparse landmarks. In Fig.\ref{subfig:e} and Fig.\ref{subfig:f}, the concentration of landmarks in the upper and lower-right parts of the image causes RIPGeo to predict based on these areas, neglecting others. HMCGeo, however, focuses on sparse landmarks in the bottom-left, correctly predicting the target host's region.
\end{itemize}

These phenomena arise from the limitations of regression methods. RIPGeo and similar models optimize parameters using losses like MSE, which struggle with noise in the input data. As a result, regression methods tend to predict the landmark center to minimize risk. This leads to biased predictions. Outlier landmarks contribute disproportionately to the gradient, shifting predictions towards them, while regions with more landmarks dominate the predictions. In contrast, HMCGeo frames IP geolocation as a classification problem, using a composite loss to focus on region correctness and adherence to hierarchical constraints, resulting in more accurate region predictions.

\subsection{Geolocation Error Analysis} \label{Geolocation Error Analysis}

To provide a more comprehensive analysis of HMCGeo in IP geolocation, we design a localization error analysis experiment. The experimental setup is as follows: we convert HMCGeo region predictions into coordinates and compare localization errors with regression methods. For example, if HMCGeo predicts a target region as \textit{``Xuhui District''}, \textit{``Lingyun Road Street''}, \textit{``Xingrongyuan Community''}, we calculate the error between the AOI center of \textit{``Xingrongyuan Community''} ($[121.4283, 31.1317]$) and the true coordinates. If multiple regions are predicted (as in HMCGeo-2 and HMCGeo-3), we compute the localization error for each region and select the result with the smallest error. The localization error CDF curves for different methods are shown in Fig.\ref{fig:Geolocation Error}.

We observe that the CDF curves for GraphGeo, RIPGeo, and TrustGeo are similar, while HMCGeo shows a clear difference from the regression methods. In the first half of the CDF curves, HMCGeo outperforms the regression methods. Specifically, for the New York and Los Angeles datasets, HMCGeo reduces the median error from 0.86 km and 1.70 km for classification methods to 0.75 km and 1.60 km, respectively. However, in the second half of the curve, HMCGeo performs worse than the regression methods. When input data contains noise, regression methods such as GraphGeo, RIPGeo, and TrustGeo may produce homogenized outputs, resulting in very similar CDF curves. HMCGeo, on the other hand, directly predicts possible region locations, avoiding homogenized predictions caused by optimizing for global accuracy, thus increasing the chances of correctly identifying the target host's region. However, when an incorrect region is predicted, it may lead to more severe prediction errors. By predicting multiple regions, this issue can be mitigated. As shown in the CDF curves for HMCGeo-2 and HMCGeo-3, these methods outperform regression methods throughout the entire process. This highlights the potential of HMCGeo in reducing IP geolocation errors.

\subsection{Sensitivity Analysis} \label{Sensitivity Analysis}
\begin{figure}[h]
    \centering
    \subfloat[\scriptsize New York\label{subfig:sensitivity_a}]{\includegraphics[width=0.16\textwidth]{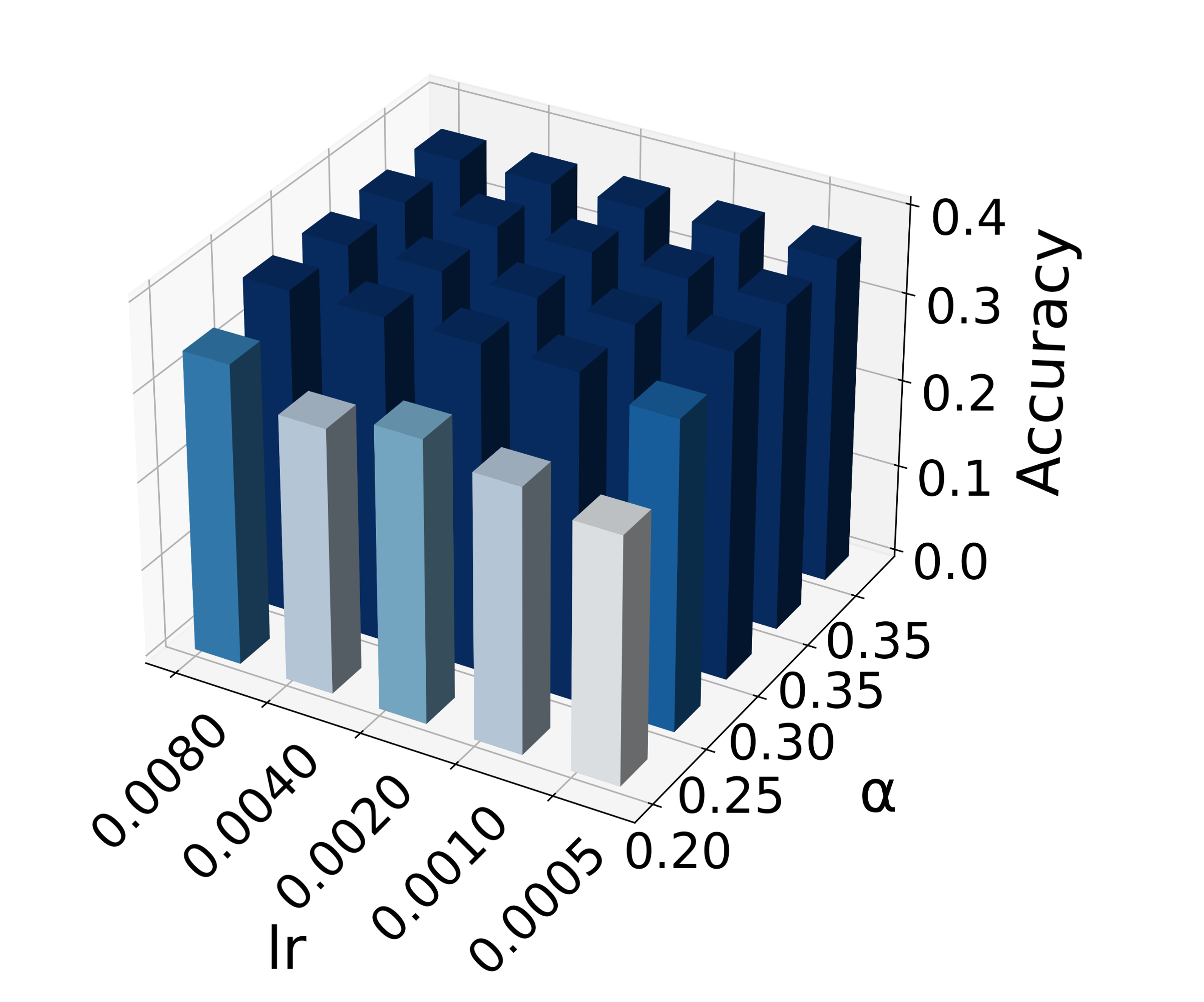}}\hfill
    \subfloat[\scriptsize Los Angeles\label{subfig:sensitivity_b}]{\includegraphics[width=0.16\textwidth]{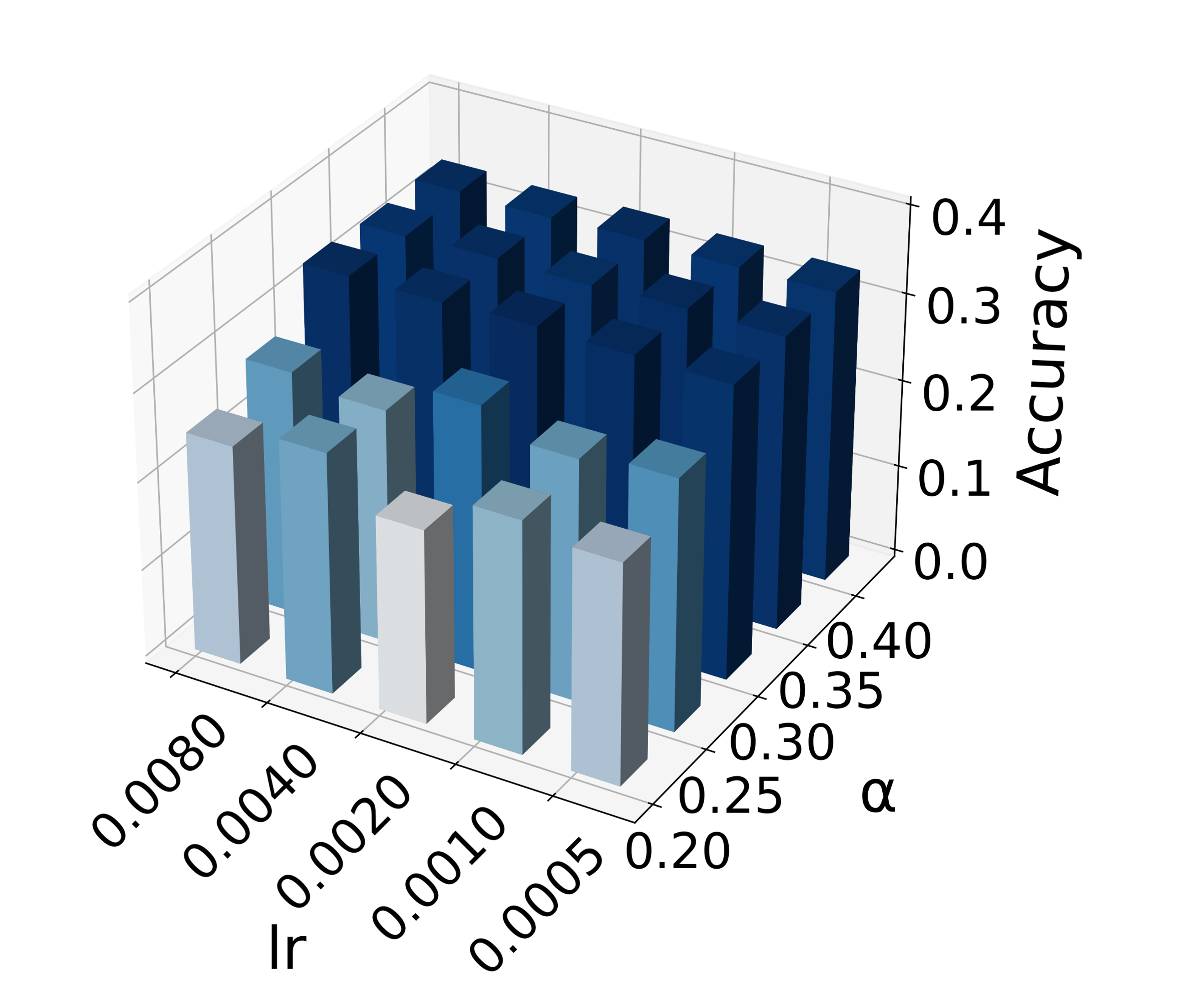}}\hfill
    \subfloat[\scriptsize Shanghai\label{subfig:sensitivity_c}]{\includegraphics[width=0.16\textwidth]{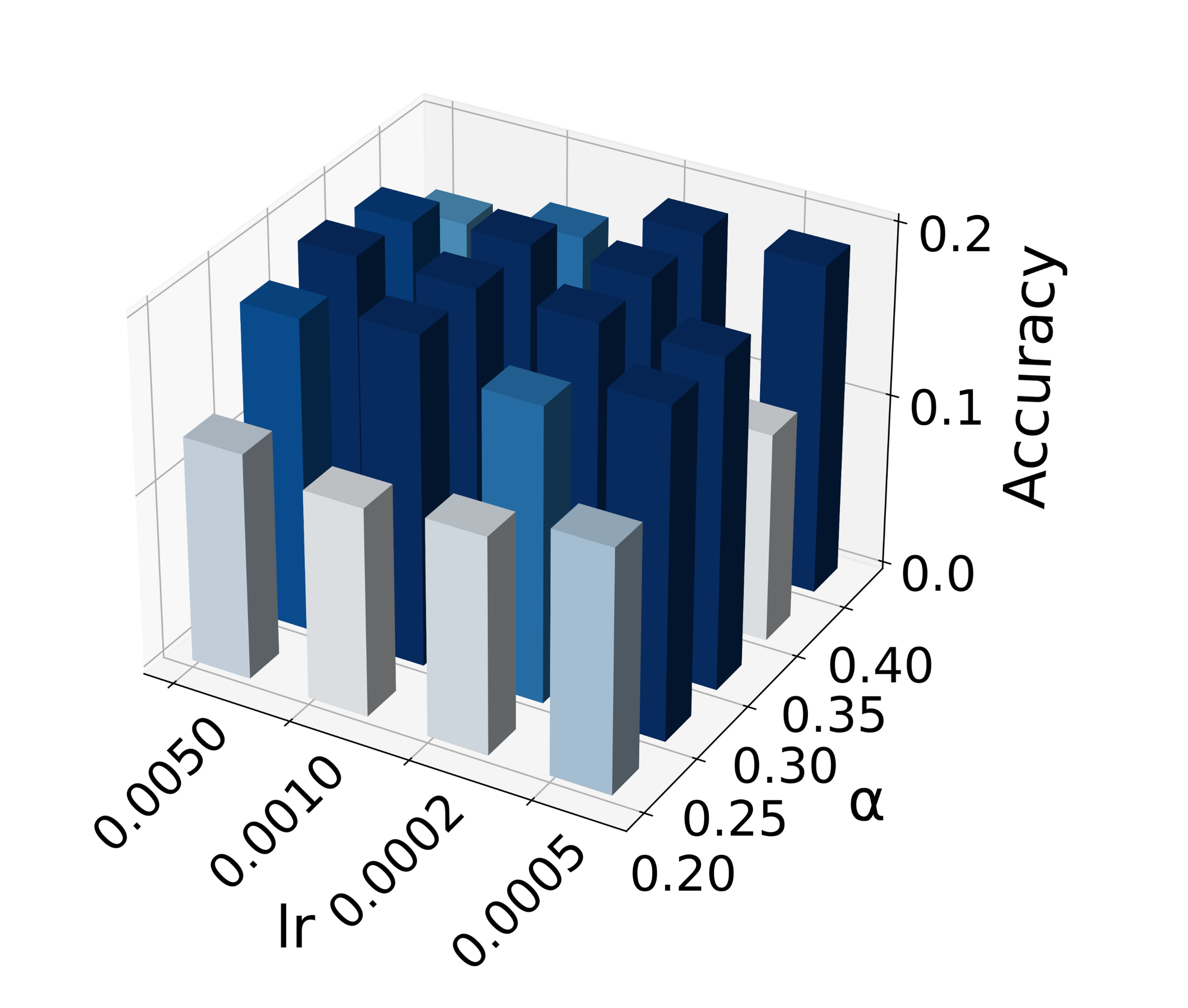}}\\
    \subfloat[\scriptsize New York\label{subfig:sensitivity_d}]{\includegraphics[width=0.16\textwidth]{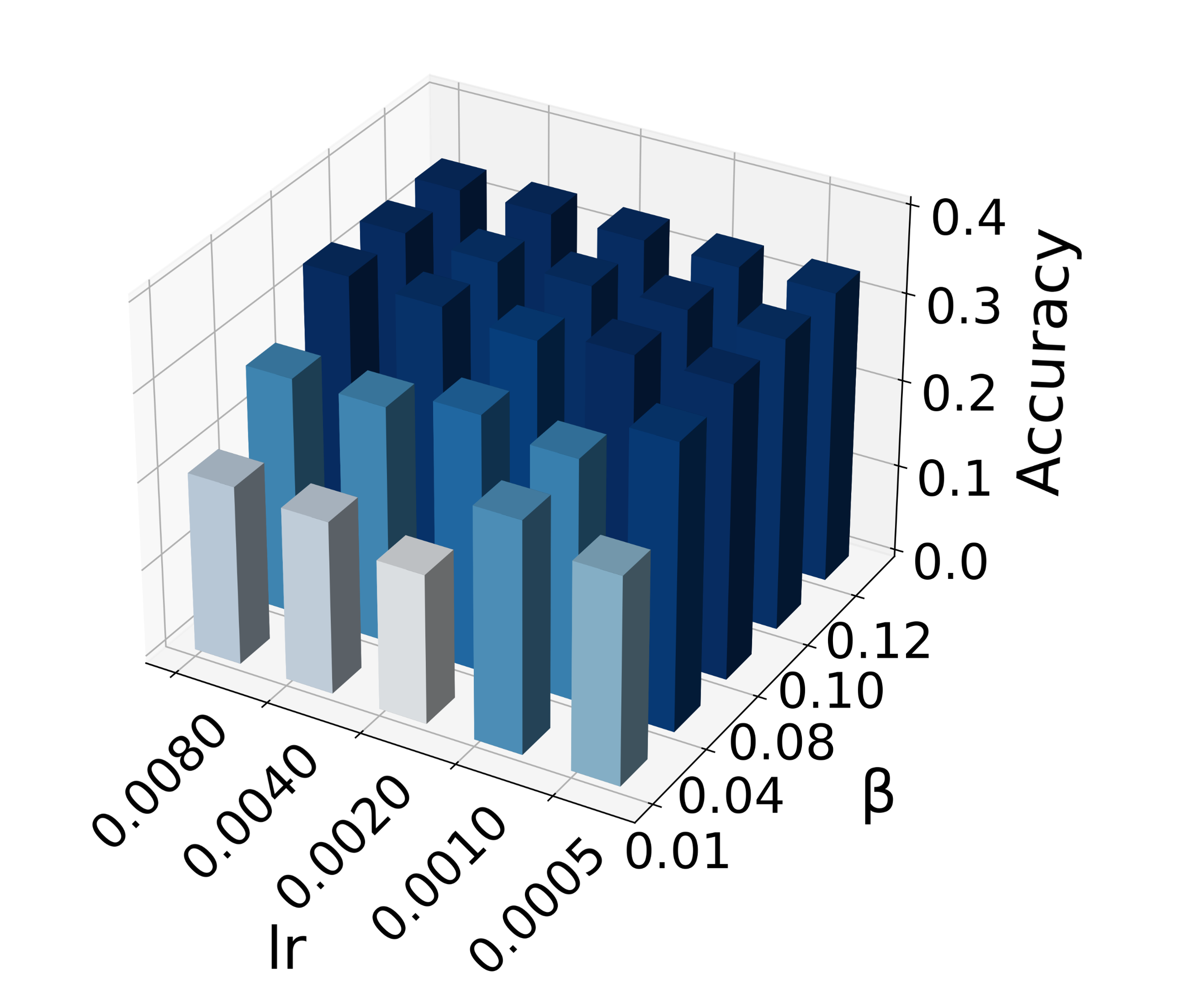}}\hfill
    \subfloat[\scriptsize Los Angeles\label{subfig:sensitivity_e}]{\includegraphics[width=0.16\textwidth]{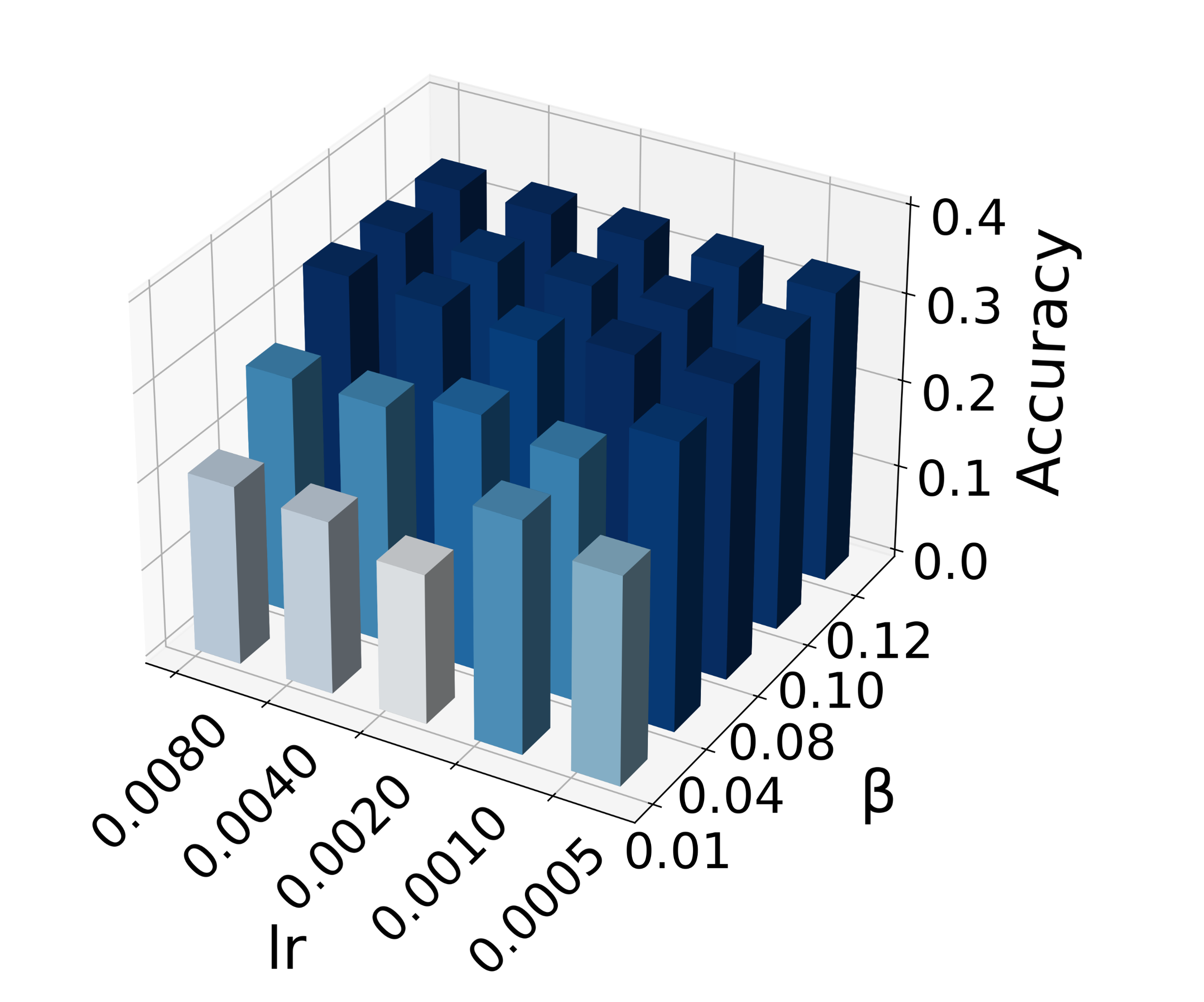}}\hfill
    \subfloat[\scriptsize Shanghai\label{subfig:sensitivity_f}]{\includegraphics[width=0.16\textwidth]{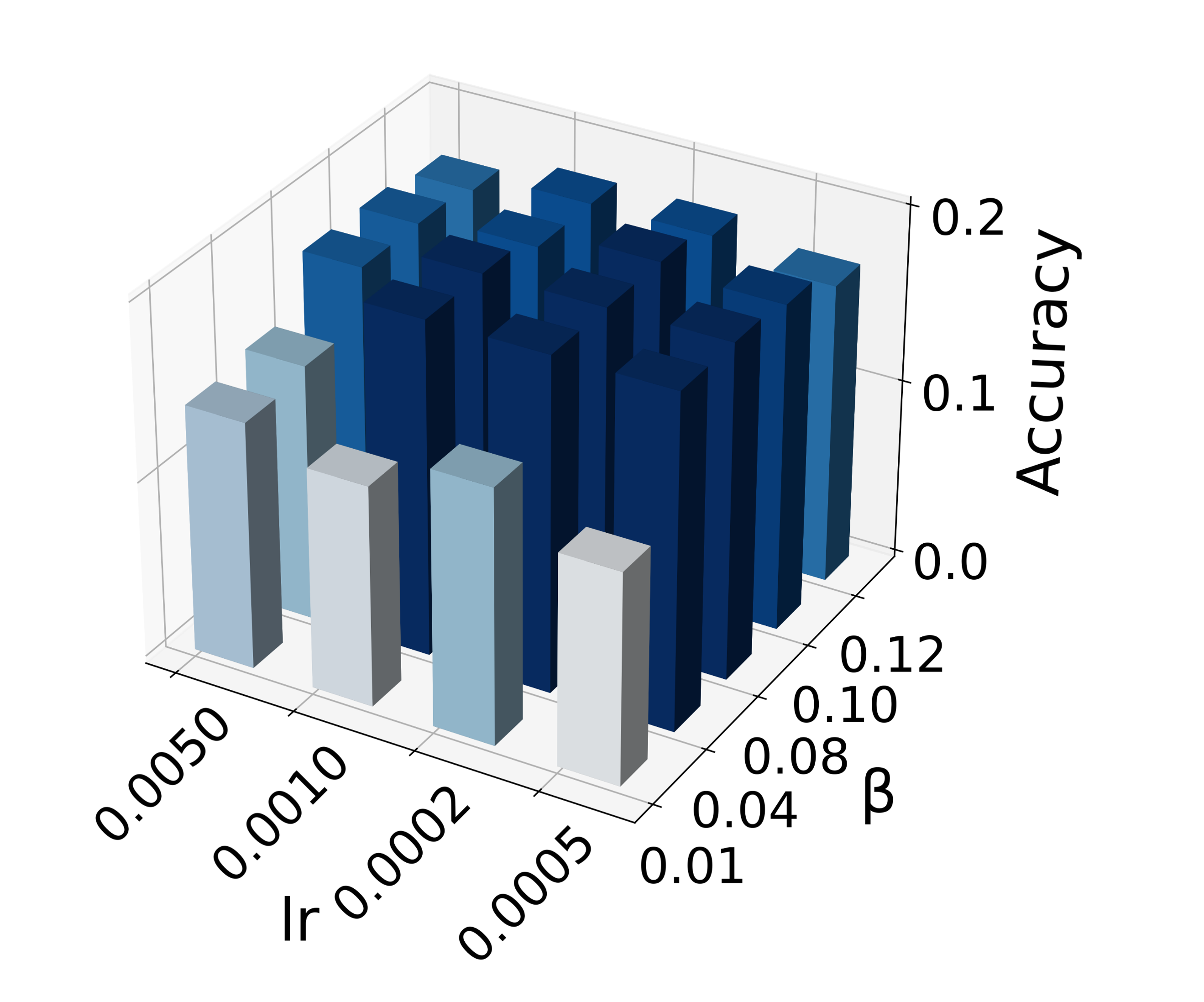}}
    \caption{Hyperparameter sensitivity analysis in HMCGeo. Figures a, b, and c analyze the impact of parameter $\alpha$, while figures d, e, and f analyze the impact of parameter $\beta$.}
\label{fig:sensitivity}
\end{figure}
We design a sensitivity analysis experiment to explore the impact of hyperparameters on HMCGeo performance, including model structure-related hyperparameters $\alpha$ and $\beta$, as well as the training-related hyperparameter $lr$. Fig.\ref{subfig:sensitivity_a}, Fig.\ref{subfig:sensitivity_b}, and Fig.\ref{subfig:sensitivity_c} show the effects of adjusting $\alpha$, while Fig.\ref{subfig:sensitivity_d}, Fig.\ref{subfig:sensitivity_e}, and Fig.\ref{subfig:sensitivity_f} illustrate the impact of adjusting $\beta$ on region prediction accuracy under different $lr$ values.

Fig.\ref{subfig:sensitivity_a}, Fig.\ref{subfig:sensitivity_b}, and Fig.\ref{subfig:sensitivity_c} show that, regardless of $lr$, accuracy increases and then decreases as $\alpha$ increases, with the optimal $\alpha$ typically being less than 0.5. Since global output is based on the finest granularity, a larger weight for global output theoretically improves predictions. However, local output focuses on single-granularity features, and a balanced combination of local and global outputs enhances overall prediction quality. As $\alpha$ increases, the global output weight decreases, leading to the observed trend. Fig.\ref{subfig:sensitivity_d}, Fig.\ref{subfig:sensitivity_e}, and Fig.\ref{subfig:sensitivity_f} show that, regardless of $lr$, accuracy follows a similar trend with $\beta$, first increasing and then decreasing, with the optimal $\beta$ usually being less than 0.5. As $\beta$ increases, the probabilistic classification loss weight increases. This suggests that a certain amount of probabilistic classification loss improves HMCGeo's region prediction, but hierarchical cross-entropy loss remains the primary factor. These results align with the ablation experiment findings: cross-entropy loss aids in accurate predictions, and a well-calibrated probabilistic classification loss further improves HMCGeo's performance by leveraging hierarchical constraints.

\section{Conclusion} \label{Conclusion}
Current IP geolocation features contain persistent noise, which results in errors at the kilometer level in fine-grained IP geolocation methods and leads to incorrect region predictions. To address these challenges, we propose HMCGeo, a framework focused on IP region prediction. Treating fine-grained IP geolocation as a classification problem, HMCGeo uses hierarchical multi-label classification to predict the target host’s region across various granularities. We divide a city into regions based on administrative boundaries, postal codes, and census blocks, then map landmark hosts to different granularities and summarize their hierarchical relationships. A topology-based landmark selection process is used to choose the appropriate landmarks for the target host. In hierarchical region prediction, both local and global outputs are computed through residual-based feature extraction and attention units. We also introduce probabilistic classification loss to enhance prediction by utilizing hierarchical relationships. Experiments on datasets from New York, Los Angeles, and Shanghai demonstrate HMCGeo's strong performance in region prediction.

While HMCGeo significantly improves region prediction, challenges remain. It struggles with data noise, which can lead to larger errors than regression methods when incorrect regions are predicted. Although selecting multiple candidates can help, it does not fully address the issue. Additionally, HMCGeo requires multiple uses of the softmax function for each granularity, reducing efficiency. Improving computational efficiency is a key area for future work.

\bibliography{references}

\begin{thebibliography}{10}
\providecommand{\url}[1]{#1}
\csname url@samestyle\endcsname
\providecommand{\newblock}{\relax}
\providecommand{\bibinfo}[2]{#2}
\providecommand{\BIBentrySTDinterwordspacing}{\spaceskip=0pt\relax}
\providecommand{\BIBentryALTinterwordstretchfactor}{4}
\providecommand{\BIBentryALTinterwordspacing}{\spaceskip=\fontdimen2\font plus
\BIBentryALTinterwordstretchfactor\fontdimen3\font minus \fontdimen4\font\relax}
\providecommand{\BIBforeignlanguage}[2]{{%
\expandafter\ifx\csname l@#1\endcsname\relax
\typeout{** WARNING: IEEEtran.bst: No hyphenation pattern has been}%
\typeout{** loaded for the language `#1'. Using the pattern for}%
\typeout{** the default language instead.}%
\else
\language=\csname l@#1\endcsname
\fi
#2}}
\providecommand{\BIBdecl}{\relax}
\BIBdecl

\bibitem{1}
J.~A. Muir and P.~C.~V. Oorschot, ``Internet geolocation: Evasion and counterevasion,'' \emph{Acm computing surveys (csur)}, vol.~42, no.~1, pp. 1--23, 2009.

\bibitem{35}
P.~Callejo, M.~Gramaglia, R.~Cuevas, and A.~Cuevas, ``A deep dive into the accuracy of ip geolocation databases and its impact on online advertising,'' \emph{IEEE Transactions on Mobile Computing}, 2022.

\bibitem{2}
C.~Guo, Y.~Liu, W.~Shen, H.~J. Wang, Q.~Yu, and Y.~Zhang, ``Mining the web and the internet for accurate ip address geolocations,'' in \emph{IEEE INFOCOM 2009}.\hskip 1em plus 0.5em minus 0.4em\relax IEEE, 2009, pp. 2841--2845.

\bibitem{3}
T.~Qiu, B.~Li, W.~Qu, E.~Ahmed, and X.~Wang, ``Tosg: A topology optimization scheme with global small world for industrial heterogeneous internet of things,'' \emph{IEEE Transactions on Industrial Informatics}, vol.~15, no.~6, pp. 3174--3184, 2018.

\bibitem{4}
M.~Gharaibeh, A.~Shah, B.~Huffaker, H.~Zhang, R.~Ensafi, and C.~Papadopoulos, ``A look at router geolocation in public and commercial databases,'' in \emph{Proceedings of the 2017 Internet Measurement Conference}, 2017, pp. 463--469.

\bibitem{5}
S.~Ding, F.~Zhao, and X.~Luo, ``A street-level ip geolocation method based on delay-distance correlation and multilayered common routers,'' \emph{Security and Communication Networks}, vol. 2021, pp. 1--11, 2021.

\bibitem{6}
Y.~Wang, D.~Burgener, M.~Flores, A.~Kuzmanovic, and C.~Huang, ``Towards street-level client-independent ip geolocation,'' in \emph{8th USENIX Symposium on Networked Systems Design and Implementation (NSDI 11)}, 2011.

\bibitem{7}
H.~Huang, G.~Gartner, J.~M. Krisp, M.~Raubal, and N.~Van~de Weghe, ``Location based services: ongoing evolution and research agenda,'' \emph{Journal of Location Based Services}, vol.~12, no.~2, pp. 63--93, 2018.

\bibitem{13}
H.~Liu, Y.~Zhang, Y.~Zhou, D.~Zhang, X.~Fu, and K.~Ramakrishnan, ``Mining checkins from location-sharing services for client-independent ip geolocation,'' in \emph{IEEE INFOCOM 2014-IEEE conference on computer communications}.\hskip 1em plus 0.5em minus 0.4em\relax IEEE, 2014, pp. 619--627.

\bibitem{14}
Q.~Li, Z.~Wang, D.~Tan, J.~Song, H.~Wang, L.~Sun, and J.~Liu, ``Geocam: An ip-based geolocation service through fine-grained and stable webcam landmarks,'' \emph{IEEE/ACM Transactions on Networking}, vol.~29, no.~4, pp. 1798--1812, 2021.

\bibitem{18}
S.~Ding, X.~Luo, J.~Wang, and X.~Fu, ``Gnn-geo: A graph neural network-based fine-grained ip geolocation framework,'' \emph{IEEE Transactions on Network Science and Engineering}, vol.~10, no.~6, pp. 3543--3560, 2023.

\bibitem{8}
Z.~Wang, F.~Zhou, W.~Zeng, G.~Trajcevski, C.~Xiao, Y.~Wang, and K.~Chen, ``Connecting the hosts: Street-level ip geolocation with graph neural networks,'' in \emph{Proceedings of the 28th ACM SIGKDD Conference on Knowledge Discovery and Data Mining}, 2022, pp. 4121--4131.

\bibitem{17}
F.~Zhang, F.~Liu, and X.~Luo, ``Geolocation of covert communication entity on the internet for post-steganalysis,'' \emph{EURASIP Journal on Image and Video Processing}, vol. 2020, pp. 1--10, 2020.

\bibitem{19}
W.~Tai, B.~Chen, F.~Zhou, T.~Zhong, G.~Trajcevski, Y.~Wang, and K.~Chen, ``Trustgeo: Uncertainty-aware dynamic graph learning for trustworthy ip geolocation,'' in \emph{Proceedings of the 29th ACM SIGKDD Conference on Knowledge Discovery and Data Mining}, 2023, pp. 4862--4871.

\bibitem{20}
W.~Tai, B.~Chen, T.~Zhong, Y.~Wang, K.~Chen, and F.~Zhou, ``Ripgeo: Robust street-level ip geolocation,'' in \emph{2023 24th IEEE International Conference on Mobile Data Management (MDM)}.\hskip 1em plus 0.5em minus 0.4em\relax IEEE, 2023, pp. 138--147.

\bibitem{9}
B.~Gueye, A.~Ziviani, M.~Crovella, and S.~Fdida, ``Constraint-based geolocation of internet hosts,'' in \emph{Proceedings of the 4th ACM SIGCOMM conference on Internet measurement}, 2004, pp. 288--293.

\bibitem{10}
E.~Katz-Bassett, J.~P. John, A.~Krishnamurthy, D.~Wetherall, T.~Anderson, and Y.~Chawathe, ``Towards ip geolocation using delay and topology measurements,'' in \emph{Proceedings of the 6th ACM SIGCOMM conference on Internet measurement}, 2006, pp. 71--84.

\bibitem{11}
V.~N. Padmanabhan and L.~Subramanian, ``An investigation of geographic mapping techniques for internet hosts,'' in \emph{Proceedings of the 2001 conference on Applications, technologies, architectures, and protocols for computer communications}, 2001, pp. 173--185.

\bibitem{12}
I.~Poese, S.~Uhlig, M.~A. Kaafar, B.~Donnet, and B.~Gueye, ``Ip geolocation databases: Unreliable?'' \emph{ACM SIGCOMM Computer Communication Review}, vol.~41, no.~2, pp. 53--56, 2011.

\bibitem{45}
{MaxMind}, ``Maxmind geoip database and services,'' \url{https://www.maxmind.com}, accessed: 2025-01-22.

\bibitem{46}
{Digital Element}, ``Digital element ip intelligence and geolocation solutions,'' \url{https://info.digitalelement.com}, accessed: 2025-01-22.

\bibitem{47}
{CZ88.net}, ``Cz88 ip database and tools,'' \url{https://cz88.net/}, accessed: 2025-01-22.

\bibitem{15}
O.~Dan, V.~Parikh, and B.~D. Davison, ``Ip geolocation through reverse dns,'' \emph{ACM Transactions on Internet Technology (TOIT)}, vol.~22, no.~1, pp. 1--29, 2021.

\bibitem{39}
------, ``Distributed reverse dns geolocation,'' in \emph{2018 IEEE International Conference on Big Data (Big Data)}.\hskip 1em plus 0.5em minus 0.4em\relax IEEE, 2018, pp. 1581--1586.

\bibitem{40}
------, ``Ip geolocation using traceroute location propagation and ip range location interpolation,'' in \emph{Companion Proceedings of the Web Conference 2021}, 2021, pp. 332--338.

\bibitem{41}
R.~Li, Y.~Liu, Y.~Qiao, T.~Ma, B.~Wang, and X.~Luo, ``Street-level landmarks acquisition based on svm classifiers.'' \emph{Computers, Materials \& Continua}, vol.~59, no.~2, 2019.

\bibitem{42}
R.~Li, R.~Xu, Y.~Ma, and X.~Luo, ``Landmarkminer: Street-level network landmarks mining method for ip geolocation,'' \emph{ACM Transactions on Internet of Things}, vol.~2, no.~3, pp. 1--22, 2021.

\bibitem{43}
J.~Lin, C.~Li, W.~Gong, G.~Song, L.~Fan, Z.~Wang, and J.~Yang, ``Probegeo: A comprehensive landmark mining framework based on web content,'' \emph{IEEE/ACM Transactions on Networking}, 2024.

\bibitem{16}
H.~Jiang, Y.~Liu, and J.~N. Matthews, ``Ip geolocation estimation using neural networks with stable landmarks,'' in \emph{2016 IEEE Conference on Computer Communications Workshops (INFOCOM WKSHPS)}.\hskip 1em plus 0.5em minus 0.4em\relax IEEE, 2016, pp. 170--175.

\bibitem{36}
N.~A. Asif, Y.~Sarker, R.~K. Chakrabortty, M.~J. Ryan, M.~H. Ahamed, D.~K. Saha, F.~R. Badal, S.~K. Das, M.~F. Ali, S.~I. Moyeen \emph{et~al.}, ``Graph neural network: A comprehensive review on non-euclidean space,'' \emph{IEEE Access}, vol.~9, pp. 60\,588--60\,606, 2021.

\bibitem{37}
S.~Zafeiriou, M.~Bronstein, T.~Cohen, O.~Vinyals, L.~Song, J.~Leskovec, P.~Li{\`o}, J.~Bruna, and M.~Gori, ``Guest editorial: Non-euclidean machine learning,'' \emph{IEEE Transactions on Pattern Analysis and Machine Intelligence}, vol.~44, no.~2, pp. 723--726, 2022.

\bibitem{44}
X.~Wang, D.~Zhao, X.~Liu, Z.~Zhang, and T.~Zhao, ``Neighborgeo: Ip geolocation based on neighbors,'' \emph{Computer Networks}, vol. 257, p. 110896, 2025.

\bibitem{21}
Y.~Meng, J.~Shen, C.~Zhang, and J.~Han, ``Weakly-supervised hierarchical text classification,'' in \emph{Proceedings of the AAAI conference on artificial intelligence}, vol.~33, no.~01, 2019, pp. 6826--6833.

\bibitem{22}
I.~Dimitrovski, D.~Kocev, S.~Loskovska, and S.~D{\v{z}}eroski, ``Hierarchical annotation of medical images,'' \emph{Pattern Recognition}, vol.~44, no. 10-11, pp. 2436--2449, 2011.

\bibitem{23}
C.~Vens, J.~Struyf, L.~Schietgat, S.~D{\v{z}}eroski, and H.~Blockeel, ``Decision trees for hierarchical multi-label classification,'' \emph{Machine learning}, vol.~73, pp. 185--214, 2008.

\bibitem{24}
W.~Bi and J.~T. Kwok, ``Multi-label classification on tree-and dag-structured hierarchies,'' in \emph{Proceedings of the 28th International Conference on International Conference on Machine Learning}, 2011, pp. 17--24.

\bibitem{25}
R.~Cerri, R.~C. Barros, and A.~C. de~Carvalho, ``A genetic algorithm for hierarchical multi-label classification,'' in \emph{Proceedings of the 27th annual ACM symposium on applied computing}, 2012, pp. 250--255.

\bibitem{26}
R.~Cerri, R.~C. Barros, and A.~C. De~Carvalho, ``Hierarchical multi-label classification using local neural networks,'' \emph{Journal of Computer and System Sciences}, vol.~80, no.~1, pp. 39--56, 2014.

\bibitem{27}
J.~Wehrmann, R.~Cerri, and R.~Barros, ``Hierarchical multi-label classification networks,'' in \emph{International conference on machine learning}.\hskip 1em plus 0.5em minus 0.4em\relax PMLR, 2018, pp. 5075--5084.

\bibitem{28}
M.~Ester, H.-P. Kriegel, J.~Sander, X.~Xu \emph{et~al.}, ``A density-based algorithm for discovering clusters in large spatial databases with noise,'' in \emph{kdd}, vol.~96, no.~34, 1996, pp. 226--231.

\bibitem{31}
\BIBentryALTinterwordspacing
{New York City Government}, ``{New York City Open Data Portal},'' 2024, [Online; accessed 26-April-2024]. [Online]. Available: \url{https://opendata.cityofnewyork.us}
\BIBentrySTDinterwordspacing

\bibitem{32}
\BIBentryALTinterwordspacing
{Los Angeles City Government}, ``{Los Angeles Open Data Portal},'' n.d., [Online; accessed 26-April-2024]. [Online]. Available: \url{https://data.lacity.org}
\BIBentrySTDinterwordspacing

\bibitem{33}
\BIBentryALTinterwordspacing
{POI86}, ``{POI86 - Point of Interest Data},'' n.d., [Online; accessed 26-April-2024]. [Online]. Available: \url{https://www.poi86.com}
\BIBentrySTDinterwordspacing

\bibitem{34}
\BIBentryALTinterwordspacing
{OpenStreetMap Contributors}, ``{OpenStreetMap},'' n.d., [Online; accessed 26-April-2024]. [Online]. Available: \url{https://www.openstreetmap.org}
\BIBentrySTDinterwordspacing

\bibitem{29}
J.~Chen, P.~Wang, J.~Liu, and Y.~Qian, ``Label relation graphs enhanced hierarchical residual network for hierarchical multi-granularity classification,'' in \emph{Proceedings of the IEEE/CVF Conference on Computer Vision and Pattern Recognition}, 2022, pp. 4858--4867.

\bibitem{30}
A.~Vaswani, N.~Shazeer, N.~Parmar, J.~Uszkoreit, L.~Jones, A.~N. Gomez, {\L}.~Kaiser, and I.~Polosukhin, ``Attention is all you need,'' \emph{Advances in neural information processing systems}, vol.~30, 2017.

\end{thebibliography}
\end{document}